\documentclass[10pt]{article} 
\usepackage[preprint]{tmlr}

\usepackage{amsmath,amsfonts,bm}

\def\eqref#1{equation~\ref{#1}}

\def\1{\bm{1}}

\DeclareMathAlphabet{\mathsfit}{\encodingdefault}{\sfdefault}{m}{sl}
\SetMathAlphabet{\mathsfit}{bold}{\encodingdefault}{\sfdefault}{bx}{n}

\usepackage{hyperref}
\usepackage{url}
\usepackage{amssymb}
\usepackage{graphicx}
\usepackage{subfigure}
\usepackage[capitalize,noabbrev]{cleveref}
\usepackage{booktabs}
\usepackage{multirow}
\usepackage[normalem]{ulem}
\usepackage[left]{lineno} 

\useunder{\uline}{\ul}{}

\title{S$^2$Transformer: Scalable Structured Transformers for Global Station Weather Forecasting}

\author{\name Hongyi Chen \email chenhongyi@stu.hit.edu.cn \\
      \addr Harbin Institute of Technology, Shenzhen, China
      \AND
      \name Xiucheng Li \email lixiucheng@hit.edu.cn \\
      \addr Harbin Institute of Technology, Shenzhen, China
      \AND
      \name Xinyang Chen \email chenxinyang@hit.edu.cn\\
      \addr Harbin Institute of Technology, Shenzhen, China 
      \AND
      \name Yun Cheng \email  yun.cheng@sdsc.ethz.ch\\
      \addr Swiss Data Science Center, Zurich, Switzerland 
      \AND
      \name Jing Li \email  li.jing@hit.edu.cn \\
      \addr Harbin Institute of Technology, Shenzhen, China 
      \AND
      \name Kehai Chen \email  chenkehai@hit.edu.cn\\
      \addr Harbin Institute of Technology, Shenzhen, China 
      \AND
      \name Liqiang Nie \email  nieliqiang@gmail.com\\
      \addr Harbin Institute of Technology, Shenzhen, China} 

\def\month{09}  
\def\year{2025} 
\def\openreview{\url{https://openreview.net/forum?id=XXXX}} 

\begin{document}

\maketitle

\begin{abstract}
Global Station Weather Forecasting (GSWF) is a key meteorological research area, critical to energy, aviation, and agriculture. Existing time series forecasting methods often ignore or unidirectionally model spatial correlation when conducting large-scale global station forecasting. This contradicts the intrinsic nature underlying observations of the global weather system, limiting forecast performance. To address this, we propose a novel Spatial Structured Attention Block in this paper. It partitions the spatial graph into a set of subgraphs and instantiates Intra-subgraph Attention to learn local spatial correlation within each subgraph, and aggregates nodes into subgraph representations for message passing among the subgraphs via Inter-subgraph Attention---considering both spatial proximity and global correlation. Building on this block, we develop a multiscale spatiotemporal forecasting model S$^2$Transformer by progressively expanding subgraph scales. The resulting model is both scalable and able to produce structured spatial correlation, and meanwhile, it is easy to implement. The experimental results show that it can achieve performance improvements up to 16.8\% over time series forecasting baselines at low running costs.
\end{abstract}

\section{Introduction}
\label{Introdution}

Global Station Weather Forecasting (GSWF) is vital to modern society, with significant implications for various sectors, including energy \citep{dehalwar2016electricity}, aviation \citep{gultepe2019review}, and agriculture \citep{ukhurebor2022precision}. Unlike regular, image-like data structures such as the Earth Reanalysis 5 \citep{hersbach2020era5}, weather station data comprises precise, fine-grained meteorological observations distributed across irregular spatial locations. Compared to radars and satellites, weather stations offer greater flexibility in acquiring scattered data and involve lower deployment costs \citep{wu2023interpretable}. Currently, physics-based Numerical Weather Prediction (NWP) models are the most widely used and reputedly the most accurate for GSWF. However, they are inherently computationally intensive, requiring vast resources during execution. Recently, Time Series Forecasting (TSF) methods, which leverage historical weather observations to predict future conditions, have demonstrated superior performance on small-scale weather station datasets and provided a more cost-effective alternative to traditional physics-based prediction models \citep{han2024far}. However, most of these methods treat GSWF as a multivariate time series forecasting task, trained and tested solely on single-station datasets. They overlook spatial correlation, contradict the intrinsic nature of the global weather system underlying observations, and thus limit forecast performance. 

A more appropriate approach is to treat GSWF as a spatiotemporal series forecasting task. Spatiotemporal data is a particular type of multivariate time series data in which each variate is equipped with a spatial coordinate in a metric space. Aside from the general topology shared by multivariate time series, spatiotemporal data also possess the spatial proximity induced by the spatial coordinates, which serves as a strong prior and has a prominent impact on the topology generation. This is summarized by Tobler's first law of geography---everything is related to everything else, but near things are more related than distant things \citep{miller2004tobler}. The spatial proximity can be described by a spatial graph, in which each variate is a node, and an edge is drawn between two nodes if their spatial distance is less than a predefined threshold. In the spirit of Tobler's first law, Airphynet \citep{DBLP:conf/iclr/HettigeJXLCW24} and Air-DualODE \citep{DBLP:conf/iclr/TianL0CGZPRY25} encode the spatial prior into the model by running graph convolution on the spatial graph to make the prediction. Despite the achievement, these methods are mostly trained and tested on meteorological station datasets with localized regions, which limits their applicability in global scenarios. More importantly, it makes distant node pairs very difficult to exchange information due to the limitations (such as over-smoothing and over-squashing) of the message-passing paradigm \citep{DBLP:conf/icml/Ma0LRDCTL23}. However, Tobler's second law of geography points out that the phenomenon external to a geographic area of interest affects what goes on inside \citep{tobler2004second}, and many works \citep{STWA, DBLP:journals/tkde/GuoLWLC22} have also confirmed that the distant nodes could also manifest strong correlation. To sidestep this, Corrformer \citep{wu2023interpretable} organizes global stations into a hierarchical tree structure, with stations under each intermediate node undergoing sequential spatial correlation modeling. Despite achieving linear complexity in modeling global spatial correlation, its unidirectional sequential approach struggles to accommodate the prevalent bidirectional or multidirectional spatial correlation in reality. Inspired by Transformer's milestones in NLP and computer vision, many spatiotemporal forecasting proposals \citep{bai2020adaptive, wu2020connecting, shang2021discrete} simply discard the spatial graph and learn spatial correlation end-to-end via an attention mechanism. The attention mechanism enables instant communication between any two nodes, effectively resolving issues with distant message passing and demonstrating significant potential in GSWF. However, such attention-based methods have two drawbacks: 1) The spatial correlation learning is unstructured (lacking structural information embedded in spatial graph) and contains numerous trivial nonzero entries (noise), the accumulated noise is likely to impair the forecasting performance when station number $N$ is large; 2) The computational and memory complexity of generating pairwise spatial correlation both reach $\mathcal{O}(N^2)$, leading to enormous computation costs for large-scale stations.

To address the two limitations, we propose a novel Spatial Structured Attention Block that respects the first law of geography while also permitting long-distance message passing in this paper. The high-level idea is to partition the spatial graph into a set of subgraphs and instantiate self-attention to learn local spatial correlation within each subgraph (Intra-subgraph Attention). To capture the global spatial correlation, we further aggregate the nodes to produce subgraph representations and exchange information among the subgraphs via self-attention again (Inter-subgraph Attention). Since the entire block is differentiable, the message passed from distant nodes can be backpropagated through Inter-subgraph Attention to their correlated nodes in a parsimonious manner. To further enable the perception of spatial structure, we encode the shortest path distance between any two nodes as a bias term in the spatial attention mechanism. Moreover, we stack the proposed Spatial Structured Attention blocks with residual and gradually increase the subgraph scales to develop our eventual GSWF model S$^2$Transformer. Such a design brings the following two appealing features: 1) our proposed method adopts the spatial graph to facilitate the perception and exchange of local spatial information, and thus it can yield sparse structure and reduce both the computational and memory burdens; 2) it permits message passing between distant node pairs in a parsimonious way, and thus it is capable of capturing global spatial correlation without incurring extra noise. To summarize:
\begin{itemize}
    \item We propose a novel Spatial Structured Attention Block for GSWF, which not only perceives spatial structure but also considers both spatial proximity and global correlation. 
    \item Building on the proposed block, we develop a multiscale GSWF model S$^2$Transformer by gradually increasing the subgraph scales. The resulting model is scalable and can produce structured spatial correlation.
    \item Our proposed method is effective yet easy to implement. We evaluate its efficacy and efficiency on global station weather datasets from medium to large sizes. It can achieve performance improvements up to 16.8\% over time series forecasting baselines while maintaining low running costs. 
\end{itemize}

\section{Related Work}
\label{Related Work}

\subsection{Spatiotemporal Series Forecasting}
\label{Spatiotemporal Series Forecasting}
Spatiotemporal series forecasting, a subfield of multivariate time series analysis, has been explored for decades. Common methods for capturing temporal dependencies include recurrent neural networks (RNNs) \citep{zhao2017lstm,lai2018modeling}, convolutional neural networks (CNNs) \citep{bai2018empirical,wu2022timesnet}, and Transformer-based models \citep{vaswani2017attention,wu2021autoformer,zhou2022fedformer,patchtst2023}. Additionally, multi-layer perceptrons (MLPs) have been applied for time series forecasting \citep{zeng2023transformers,challu2023nhits,wang2024timemixer}, showing that even simple models can effectively extract strong temporal periodic patterns. Beyond temporal dependencies, spatial correlation is equally critical in spatiotemporal forecasting. The advancement of graph neural networks (GNNs) offers an effective way to model unstructured spatial adjacency correlation. In the spirit of Tobler’s first law of geography, DCRNN \citep{li2017diffusion} and TGCN \citep{zhao2019t} leverage a spatial graph based on real-world distance and propose to fuse the local spatial information via graph convolution operation. However, it makes the distant node pairs very hard to exchange information due to the limitations of the message-passing paradigm~\citep{DBLP:conf/icml/Ma0LRDCTL23}, violating Tobler’s second law of geography. Subsequently, adaptive GNN-based methods have been proposed to solve this problem. AGCRN \citep{bai2020adaptive} and  MTGNN \citep{wu2020connecting} learn a representation for each series and then generate the correlation graph via pairwise node interactions. GTS \citep{shang2021discrete} and STEP \citep{shao2022pre} directly learn a discrete graph based on historical time series. Benefiting from naturally constructing a fully connected graph with learnable edge weights, the self-attention mechanism is also a commonly adopted method for capturing global and dynamic spatial correlation \citep{jiang2023pdformer,liu2024itransformer,wang2024card}. Nevertheless, the learned spatial correlation matrix in such methods is unstructured (lacking structural information
embedded in the spatial graph) and contains a large fraction of trivial nonzero entries (noise). The accumulated noise is likely to impair the forecasting performance when \(N\) is large. Moreover, they require \(\mathcal{O}(N^2)\) computational complexity, impeding their application in large-scale datasets. Several researchers have developed scalable spatiotemporal forecasting methods to accommodate larger datasets. Detailed related work on this aspect is provided in Appendix~\ref{Scalable Spatiotemporal Series Forecasting}. 
\subsection{Data-driven Numerical Weather Prediction}
\label{Data-driven Numerical Weather Prediction}

In recent years, data-driven Numerical Weather Prediction (NWP) models based on machine learning have developed rapidly. Models including Pangu-Weather \citep{DBLP:journals/nature/BiXZCG023}, GraphCast \citep{lam2023learning}, and Aurora \citep{DBLP:journals/nature/BodnarBLSABGRWD25} have demonstrated the ability to surpass conventional physics-based NWP models in terms of forecast accuracy and operational effectiveness. However, as they operate on grid spaces, they may not be optimal for global station weather forecasting. A direct method is to treat GSWF as an independent time series forecasting task and predict meteorological factors for each station individually \citep{DBLP:journals/nn/KarevanS20, hewage2020temporal,wu2021autoformer}. However, global weather constitutes an integrated system with multi-scale interactions. These methods overlook spatial correlation, contradict the intrinsic nature of the global weather system underlying observations, and thus limit forecast performance. To address this issue, Airphynet \citep{DBLP:conf/iclr/HettigeJXLCW24} and Air-DualODE \citep{DBLP:conf/iclr/TianL0CGZPRY25} encode the prior into the model by running graph convolution on the pre-defined spatial graph. Nevertheless, these methods are mostly trained and tested on meteorological station datasets with localized regions. Learning from localized regional data often fails to capture broader spatial patterns. Furthermore, models overfitted to specific regions tend to lack generalization capability, limiting their applicability in real-world scenarios. To accommodate global station weather data, Corrformer \citep{wu2023interpretable} organizes global stations into a hierarchical tree structure, with stations under each intermediate node undergoing sequential spatial correlation modeling. Despite achieving linear complexity in modeling global spatial correlation, its unidirectional sequential approach struggles to accommodate the prevalent bidirectional or multidirectional spatial correlation in reality.

\section{Preliminaries}
\label{Preliminaries}

Global station weather data refers to multivariate time series data in which each series is associated with a station's spatial coordinates. Specifically, we define the spatial coordinates of all stations as $\boldsymbol \lambda \in \mathbb R^{N}$, $\boldsymbol \phi \in \mathbb R^{N}$ and the multivariate time series \(\mathbf{X} \in \mathbb{R}^{N \times T \times C}\) records \(C\)-dimensional meteorological variables collected by \(N\) weather stations over \(T\) time steps. With a given threshold $\epsilon$, the spatial coordinates of these stations induce a spatial graph \(G = (V, E)\), where each node corresponds to a weather station (i.e., $|V| = N$) and two nodes are connected by an edge $e \in E$ if their spatial distance is smaller than $\epsilon$. Besides, we use \(\mathbf A \in \mathbb{R}^{N\times N}\) to represent the adjacent matrix of $G$. Given the past \(T\) steps historical observations \(\mathbf X_{t-T+1:t}\), along with spatial coordinates $\boldsymbol \lambda$, $\boldsymbol \phi$ and the spatial adjacent matrix \(\mathbf A\), the goal of global station weather forecasting is to predict the future \(F\) steps of meteorological variables \(\hat{\mathbf X}_{t+1:t+F}\),
\begin{equation}
\hat{\mathbf X}_{t + 1:t+F}=\mathcal{F}_{\theta}(\mathbf X_{t - T + 1:t},\boldsymbol \lambda,\boldsymbol \phi, \mathbf A)
\end{equation}
where \(\mathcal{F}_{\theta}(\cdot)\) denotes the data-driven weather forecasting model parameterized by \(\theta\).

\section{Methodology}
\label{Methodology}

\begin{figure}
    \centering
    \includegraphics[width=0.9\textwidth]{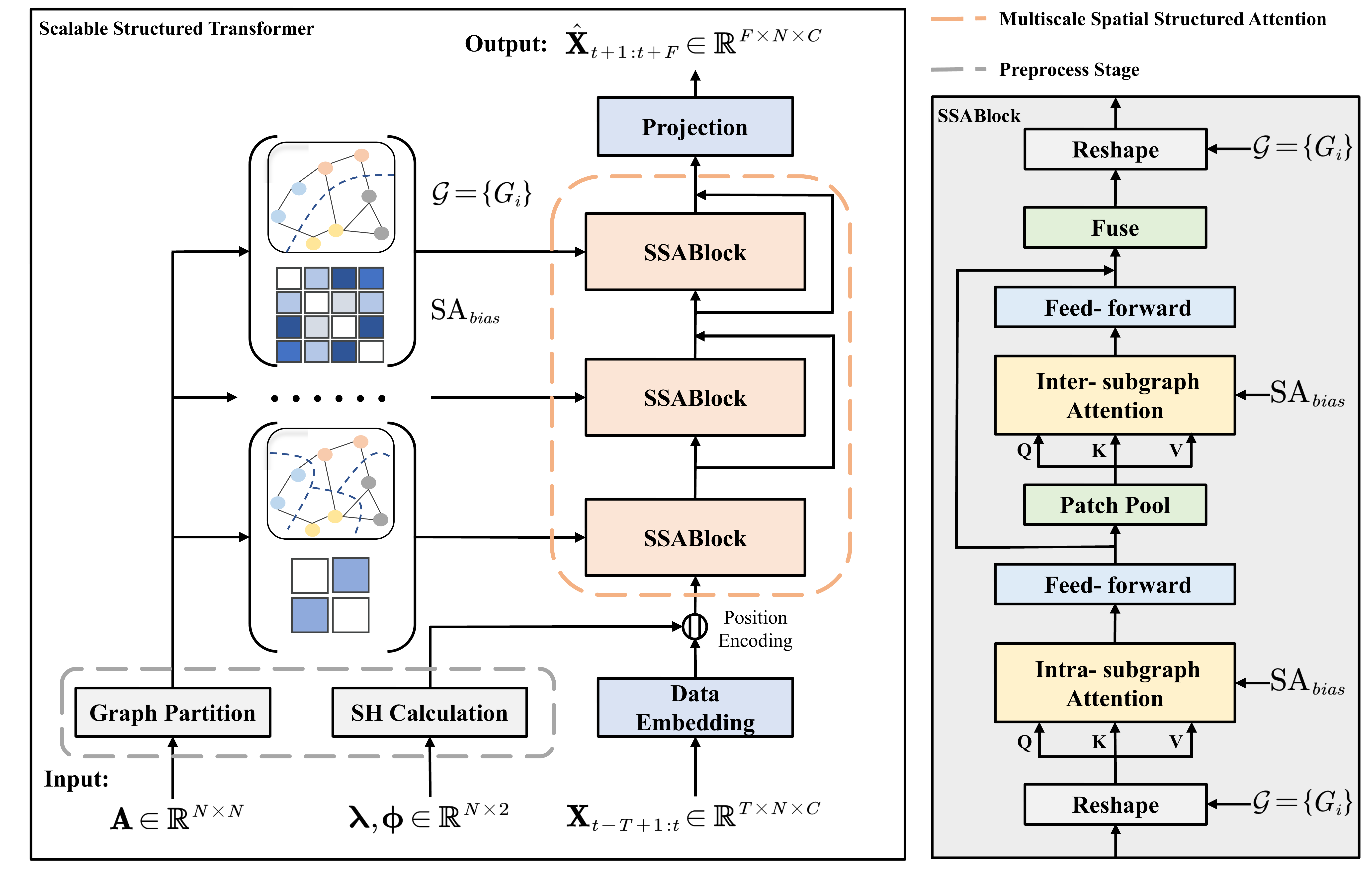}
    \caption{Left: The framework of our proposed Scalable Structured Transformer. Right: The structure of the proposed Spatial Structured Attention Block.}
    \label{fig:model}
\end{figure} 
The framework of our proposed forecasting model, dubbed Scalable Structured Transformer (S$^2$Transformer), is shown on the left of~\cref{fig:model}. We first perform spatial graph partitioning for global meteorological stations and calculate their spherical harmonics to extract spatial structure information (Section~\ref{Spatial Information Preprocess}). Following the setting of previous work \citep{liu2024itransformer,DBLP:conf/kdd/0001L0S0LJ025}, we reshape and transform the historical observation $\mathbf X_{t-T+1:t} \in \mathbb R^{N \times T \times C}$ into the input embeddings $\mathbf X_\mathrm{emb} \in \mathbb R^{N \times D}$ with a linear map, where $D$ is the embedding dimension. Next, the embedding and graph partition are fed into the Spatial Structured Attention Block (as described in the right of~\cref{fig:model} and Section~\ref{Spatial Structured Attention Block}) to fuse the information and form the contextual representations, which will be used to make the forecasting. Building on the proposed blocks, we further develop a multi-scale attention mechanism to enhance representation learning by stacking multiple Spatial Structured Attention blocks with various spatial scales (Section~\ref{Multiscale Spatial Structured Architecture}).

\subsection{Spatial Information Preprocess}
\label{Spatial Information Preprocess} 

In the preprocessing stage, we focus on extracting spatial structure information from input metadata, including spatial graph partitioning and spherical harmonics calculation. Notably, this stage does not introduce scalability bottlenecks. Compared to the training duration, the time spent on preprocessing is negligible. Detailed efficiency analysis is provided in Section~\ref{Efficiency Study}.

\textbf{Graph Partition}. Our key observation is that global meteorological observations often exhibit stronger correlation within local areas, which is also supported by Tobler's first law of geography. This motivates us to partition the spatial graph $G$ into a set of subgraphs $\mathcal G=\{G_{i}\}_{i=1}^P$, where \(G_{i} = \{V_{i}, E_{i}\}\) represents a subgraph of \(G\),  satisfying $\bigcup G_i=G$ and $\bigcap G_i=\varnothing$. Then, we can fuse the information locally in each subgraph with self-attention. For graph partitioning algorithms, we opt for the METIS algorithm \citep{karypis1998fast} given its efficiency and balanced subgraph outputs. 

\textbf{Location Embedding}. In addition, location information serves as valuable metadata in numerous geospatial applications, including GSWF. While sine-cosine embedding methods have proven effective in transformers, they assume a rectangular domain for longitude and latitude coordinates, which fails to capture the Earth’s spherical geometry accurately. Inspired by Geographic Location Encoder \citep{DBLP:conf/iclr/RusswurmKRZT24}, we employ spherical harmonic basis functions as positional embeddings, which are well-defined globally (including the poles) and enable better discrimination of weather stations distributed across the globe. These spherical harmonics are precomputed from coordinates, with their weights learned directly. Specifically, we define all station location embedding as:
\begin{equation}
    \mathrm{SH}(\boldsymbol{\lambda, \phi}) = \oplus_{n=0}^{N} \|_{l=0}^{\infty} \|_{m=-l}^{l} w^m_l {Y^m_l}({\lambda_n, \phi_n})
    \label{eq:sh}
\end{equation}
\begin{equation}
   \mathbf X_\mathrm{emb} = \mathbf X_\mathrm{emb} \| \mathrm{SH}(\boldsymbol{\lambda, \phi})
\end{equation}
where $\oplus$ indicates the stack operator and $\|$ indicates the concatenation operator. $w^m_l$ is a learnable weight shared across all stations and ${Y_l^m}$ is an orthogonal spherical harmonic basis function with increasingly higher-frequency degrees $l$ and orders $m$. In practice, we choose a maximum number $L$ instead of $\infty$ in Eq.~\ref{eq:sh}. A detailed introduction to spherical harmonics is provided in Appendix~\ref{Spherical Harmonics}.

\subsection{Spatial Structured Attention Block}
\label{Spatial Structured Attention Block} 
In this section, we introduce details of the Spatial Structured Attention (SSA) Block, guided by Tobler’s first and second laws of geography. Firstly, we reshape the input \(\mathbf{X}_\mathrm{emb}\) from \(\mathbb{R}^{N \times D}\) to \(\mathbb{R}^{P \times M \times D}\) according to the graph partition result $\mathcal G$, where \(P\) is the number of subgraphs, and \(M\) is the number of nodes in the largest subgraph. If the number of nodes in the subgraph is less than \(M\), we pad it with zeros and mask it in subsequent attention operations.

\textbf{Intra-subgraph Attention}. In light of Tobler’s first law of geography \citep{miller2004tobler}, we first employ intra-subgraph attention to model the local spatial correlation. Formally, let \(\mathbf{X}\) be the input of the block (e.g., $\mathbf{X}_\mathrm{emb}$) and \(\mathbf{X}_{p} \in \mathbb{R}^{M \times D}\) denote the embedding in the $p$-th subgraph, the representations within the subgraph are updated as follows.
\begin{align}
    \boldsymbol{\alpha}_p &= \mathrm{softmax}\left( \frac{\mathbf{Q}_{p} \mathbf{K}_{p}^{\top}} {\sqrt{d \,}} \right)  \label{eq:a-p}\\
    \mathbf{Y}_{p} &= \mathrm{FFN}\left( \boldsymbol{\alpha}_p \mathbf{V}_{p} \right)
\end{align}
\begin{equation}
    \mathbf{Q}_p, \mathbf{K}_p, \mathbf{V}_p = \left(\mathbf{W}_{Q},\mathbf{W}_{K},\mathbf{W}_{V}\right) \mathbf{X}_p
\end{equation}
where \(\mathbf{W}_{Q}\), \(\mathbf{W}_{K}\) and \(\mathbf{W}_{V}\) are learnable parameters that transform $\mathbf X_p$ into different semantic spaces, \(\boldsymbol{\alpha}_p \in \mathbb{R}^{M \times M}\) is the intra-subgraph attention map that captures the local spatial correlation in the $p$-th subgraph, and $\mathbf Y_p$ is the learned contextual representations that will be fed to the subsequent inter-subgraph attention.

\textbf{Inter-subgraph Attention}. Tobler’s second law of geography \citep{tobler2004second} states that the phenomenon external to a geographic area of interest affects what goes on inside. However, as mentioned in Section~\ref{Introdution}, directly using the attention mechanism will lead to quadratic complexity $\mathcal O(N^2)$ and incur extra noise that impairs the forecasting performance. Thus, we innovatively learn the attention between different subgraphs to approximate the global attention mechanism. We first apply mean pooling to \(\mathbf{Y}_p\) to obtain the subgraph representation \(\mathbf{s}_{p} \in \mathbb{R}^{D}\) and then stack them to produce \( \mathbf{S} \triangleq [\mathbf s_1; \mathbf s_2; \ldots ;\mathbf s_P] \in \mathbb{R}^{P \times D}\). Then, we employ the inter-subgraph attention to exchange information across subgraphs as follows.
\begin{align}
    \boldsymbol{\alpha}^\prime &= \mathrm{softmax}\left( \frac{\mathbf{Q}^\prime \mathbf{K}^{\prime \top}} {\sqrt{d \,}} \right)  \label{eq:s-p} \\
    \mathbf{S}^\prime &= \mathrm{FFN}\left( \boldsymbol{\alpha}^\prime \mathbf{V}^\prime \right) \label{eq:s-prime}
\end{align}
\begin{equation}
    \mathbf{Q}^\prime, \mathbf{K}^\prime, \mathbf{V}^\prime = \left(\mathbf{W}_{Q^\prime},\mathbf{W}_{K^\prime},\mathbf{W}_{V^\prime}\right) \mathbf{S}
\end{equation}
where \(\mathbf{W}_{Q'}\), \(\mathbf{W}_{K'}\), and \(\mathbf{W}_{V'}\) are learnable parameters. \(\boldsymbol{\alpha}^\prime \in \mathbb{R}^{P \times P}\) captures the global spatial correlation among subgraphs, which allows us to approximate the global attention mechanism by assigning the same attention value $\alpha_{pq}$ to any pair of nodes from the subgraphs $p$ and $q$. This can be considered as a sort of \emph{implicit regularization} that encourages the model to focus on mining groups of nodes that share similar patterns while ignoring unrelated noise. In other words, we only permit the distant node pairs to exchange information in a parsimonious manner, and only valuable information can be transmitted across groups. 

In the end, to take both local and global spatial information into account, we expand the shape of \(\mathbf{S}^{\prime}\) in Eq.~\ref{eq:s-prime} from \(\mathbb{R}^{P \times D}\) to \(\mathbb{R}^{P \times M \times D}\), which is concatenated with the local representation and then transformed by a linear map to produce the output:
\begin{equation}
     \mathbf{{X}}^\prime = \mathbf{W} (\mathbf{Y} \| \mathbf {S}^\prime)
\end{equation} 
where \(\|\) indicates the concatenation operator and \(\mathbf{W} \in \mathbb{R}^{2D \times D}\) is the parameter of linear map layer. 

\textbf{Spatial Attention Bias}. Self-attention serves as the core computational module in our proposed block, however, it is oblivious to the local graph structures due to its permutation invariant property. To encode the structural information into the attention mechanism, we encode the shortest path distance between any two nodes as a bias in spatial attention inspired by Graphormer \citep{DBLP:conf/nips/YingCLZKHSL21}. Specifically, we precompute the shortest path distance (SPD) between two stations. For unconnected stations, the SPD is set to a special value of $-1$. We calculate the spatial attention bias matrix $\mathbf{SA^{bias}}$ by element-wise embedding the SPD matrix with a learnable scalar. We then revise Eq.~\ref{eq:a-p} as follows:
\begin{equation}
    \boldsymbol{\alpha}_p = \mathrm{softmax}\left( \frac{\mathbf{Q}_{p} \mathbf{K}_{p}^{\top}} {\sqrt{d \,}} + \mathbf{SA}^{\mathrm{bias}}_{p} \right)
\end{equation}
\begin{equation}
    \mathbf{S A}^\mathrm{bias}_{p} = \sigma\left(\mathrm{SPD}\left(\mathbf A_p\right)\right)
\end{equation}
where $\mathbf{S A}^\mathrm{bias}_{p} \in \mathbb{R}^{M \times M}$ is the intra-subgraph attention bias matrix in the $p$-th subgraph and $\sigma$ is an element-wise learnable scalar shared across all blocks. We can modify Eq.~\ref{eq:s-p} in the same manner.


\subsection{Multiscale Spatial Structured Architecture}
\label{Multiscale Spatial Structured Architecture}

In practice, the spatial correlation often presents multiscale structures due to the multiscale property of underlying physical dynamics. To capture the intrinsic multiscale property, we develop a multiscale spatial balance architecture by stacking $L$ SSA blocks by gradually increasing the subgraph scales. Specifically, we perform graph partition with various subgraph scales to produce $L$ sets, $\mathcal G_1, \mathcal G_2, \ldots, \mathcal G_L$ such that $|\mathcal G_i| = |\mathcal G_{i-1}| / 2$. $L$ is set to $2$ by default, and we empirically find that it performs quite well in practice. Such a design also brings two additional benefits: 1) it progressively expands the receptive field of the node attention mechanism, which aligns with the spatial diffusion process of the global weather system's physical dynamics; 2) it offers the chance for spatially closed nodes at the boundary of two subgraphs to exchange information in the high-level block.  

\textbf{Forecasting}. We reshape the output of \(L\)-th block \( \mathbf{X}^{\prime(L)} \) from \(\mathbb{R}^{P \times M \times D}\) to \(\mathbb{R}^{N \times D} \) according to the graph partition result and produce the multi-step prediction \(\hat{\mathbf X}_{t_0:t_0 + F}\) through a linear projection. The model is optimized by minimizing the mean absolute error:
\begin{equation}
\mathcal{L}(\mathbf X_{t_0:t_0 + F}, \hat{\mathbf X}_{t_0:t_0 + F}) = \frac{\sum_{n = 1}^{N} \sum_{t = t_0}^{t_0 + F - 1}  \sum_{c = 0}^{C} |\hat{x}_{n,t,c} - x_{n,t,c}|}{N \times F \times C}
\end{equation}

\textbf{Complexity Analysis}. Our method achieves significant computational efficiency improvements by focusing on Intra-subgraph and Inter-subgraph Attention. Intra-subgraph Attention (for subgraphs with \(\frac{N}{P}\) nodes) has complexity \(\mathcal{O}(P(\frac{N}{P})^2D) = \mathcal{O}(\frac{N^2}{P}D)\), and Inter-subgraph Attention (for cross-subgraph correlation) has \(\mathcal{O}(P^2D)\), leading to an overall complexity of \(\mathcal{O}(\frac{N^2}{P}D + P^2D)\). Minimizing this (via derivative calculation) gives \(\mathcal{O}(2N^{4/3}D)\) when \(P = N^{2/3}\), with better scalability than quadratic-complexity methods. Our model’s efficiency can be further enhanced by linear attention; though Corrformer (specifically designed for GSWF tasks) has a better theoretical complexity of \(\mathcal{O}(NT \log T D)\), it performs poorly in practice. This is due to its temporal alignment/rearrangement operations (for inferred node order), which disrupt data layout, increase memory latency (hindering hardware parallelism), and raise memory occupancy (via intermediate data storage). Detailed efficiency analysis is provided in Section~\ref{Efficiency Study}.

\section{Experiments}

\begin{table}[]
\caption{Dataset statistics.}
\label{table:dataset-table}
\begin{center}
\begin{small}
\begin{tabular}{lllll}
\toprule
Dataset     & Frequency & Time Span & Stations & Variables Name                                   \\ \midrule
WEATHER-5K  & 1 hour    & 2014-2023 & 5672     & Wind, Temp                                       \\ \midrule
NCEI Global & 1 hour    & 2019-2020 & 3850     & Temp, Dewpoint, Wind Rate, Wind Direc, Sea Level \\ \bottomrule
\end{tabular}
\end{small}
\end{center}
\end{table}

In this section, we evaluate our approach using two benchmark datasets (Section~\ref{Performance Comparison} and Appendix~\ref{Long term Forecasting Result}). Section~\ref{Efficiency Study} presents the efficiency analysis, and Section~\ref{Ablation Study} describes the ablation study. The sensitivity of hyperparameters is detailed in Section~\ref{Parameter Sensitivity Analysis} and Appendix~\ref{More Parameters Sensitivity Analysis}. To gain a more profound understanding of our model, we also conducted visualizations, which are included in Section~\ref{Visualization} and Appendix~\ref{More Visualization}. 

\subsection{Experimental Setup}
\textbf{Datasets}. We evaluate the performance and efficiency of the proposed method on two global station weather forecasting benchmarks. The first benchmark is the WEATHER-5K dataset \citep{han2024far}, which includes crucial weather elements collected from 5672 global weather stations over ten years. The second benchmark is the NCEI Global dataset \citep{wu2023interpretable}, which contains the hourly averaged wind speed and hourly temperature of 3,850 stations worldwide from 2019 to 2020. Dataset statistics are presented in Table~\ref{table:dataset-table}, and further particulars are available in Appendix~\ref{Implementation Details}.

\textbf{Baselines}. We compare our method with the following baselines: (1) Physics-based NWP model: ECMWF-HRES \citep{EC} for WEATHER-5K dataset and ERA5 (reanalysis, 0.25°) \citep{hersbach2020era5} for NCEI Global dataset; (2) Pure time dependencies modeling methods: Informer \citep{zhou2021informer}, Autoformer \citep{wu2021autoformer}, Pyraformer \citep{DBLP:conf/iclr/LiuYLLLLD22}, STID \citep{DBLP:conf/cikm/ShaoZ00X22}; (3) Spatial correlation modeling methods: MTGNN \citep{wu2020connecting}, Corrformer \citep{wu2023interpretable}, iTransformer \citep{liu2024itransformer}; (4) Scalable spatial correlation modeling methods: RPMixer \cite{yeh2024rpmixer}, PatchSTG \citep{DBLP:conf/kdd/0001L0S0LJ025}. Notably, Pyraformer and Corrformer are the Best Time Series Forecasting methods reported on the WEATHER-5K and NCEI Global datasets. More details of baselines are provided in Appendix~\ref{Implementation Details}.

\textbf{Evaluation Metrics}. We conduct a comprehensive comparison using various evaluation criteria from the performance and efficiency perspectives. We evaluated performance using the mean absolute error (MAE) and mean square error (MSE). We consider efficiency by measuring both the training wall-clock time and maximum memory usage during training. 

\textbf{Implementation details}. Given our focus on short-term global station weather forecasting, we predict one day into the future using the past two days of data, where the input length is 48 (hours) and the predicted length is 24 (hours). The key parameter settings are detailed in Appendix~\ref{Implementation Details}. All experiments in this study are implemented using PyTorch~\cite{paszke2019pytorch} and conducted on an NVIDIA RTX 4090 GPU with 24GB memory. We run each experiment three times and report the average results.
 
\subsection{Performance Comparison}
\label{Performance Comparison}

\begin{table*}[]
\renewcommand{\arraystretch}{1.5}
\caption{Global station weather forecasting performance comparison. The best results are highlighted in \textbf{bold}, while the second-best results are \uline{underlined}.} 
\label{table:performance-comparison}
\begin{center}
\begin{small}
\resizebox{\linewidth}{!}{
\begin{tabular}{l|cccccccccc|cccc}
\hline
Dataset      & \multicolumn{10}{c|}{WEATHER-5K}                                                                                                                                                                                                                     & \multicolumn{4}{c}{NCEI Global}                                                    \\ \toprule
Variable     & \multicolumn{2}{c|}{Temperature}                   & \multicolumn{2}{c|}{Dewpoint}                      & \multicolumn{2}{c|}{Wind Rate}                     & \multicolumn{2}{c|}{Wind Direc.}                     & \multicolumn{2}{c|}{Sea Level} & \multicolumn{2}{c|}{Wind}                          & \multicolumn{2}{c}{Temp}      \\ \midrule
Metric       & MAE           & \multicolumn{1}{c|}{MSE}           & MAE           & \multicolumn{1}{c|}{MSE}           & MAE           & \multicolumn{1}{c|}{MSE}           & MAE           & \multicolumn{1}{c|}{MSE}             & MAE            & MSE           & MAE           & \multicolumn{1}{c|}{MSE}           & MAE           & MSE           \\ \midrule
NWP Model    & 1.76          & \multicolumn{1}{c|}{7.39}          & 1.85          & \multicolumn{1}{c|}{7.94}          & 1.48          & \multicolumn{1}{c|}{4.53}          & 63.8          & \multicolumn{1}{c|}{7158.3}          & \textbf{0.86}  & \textbf{2.68} & 1.59          & \multicolumn{1}{c|}{5.00}          & 1.91          & 13.45         \\
Informer     & 1.88          & \multicolumn{1}{c|}{7.51}          & 1.94          & \multicolumn{1}{c|}{8.30}          & 1.30          & \multicolumn{1}{c|}{3.62}          & 60.7          & \multicolumn{1}{c|}{6906.9}          & 2.01           & 10.56         & 1.58          & \multicolumn{1}{c|}{4.93}          & 4.42          & 33.29         \\
Autoformer   & 1.93          & \multicolumn{1}{c|}{8.64}          & 2.06          & \multicolumn{1}{c|}{9.57}          & 1.42          & \multicolumn{1}{c|}{3.97}          & 66.5          & \multicolumn{1}{c|}{7710.0}          & 2.26           & 12.78         & 1.47          & \multicolumn{1}{c|}{4.69}          & 2.25          & 10.14         \\
Pyraformer   & 1.75          & \multicolumn{1}{c|}{6.92}          & 1.83          & \multicolumn{1}{c|}{7.88}          & 1.30          & \multicolumn{1}{c|}{3.58}          & 61.8          & \multicolumn{1}{c|}{6930.2}          & 1.90           & 9.72          & 1.51          & \multicolumn{1}{c|}{4.61}          & 3.67          & 23.33         \\
STID         & 1.78          & \multicolumn{1}{c|}{7.09}          & 1.83          & \multicolumn{1}{c|}{7.88}          & 1.28          & \multicolumn{1}{c|}{3.53}          & 60.9          & \multicolumn{1}{c|}{6722.2}          & 1.87           & 9.42          & 1.34    & \multicolumn{1}{c|}{\ul 3.83}          & 1.99          & 8.46          \\
MTGNN        & 1.84          & \multicolumn{1}{c|}{7.36}          & 1.89          & \multicolumn{1}{c|}{8.18}          & 1.30          & \multicolumn{1}{c|}{3.59}          & 62.1          & \multicolumn{1}{c|}{6854.5}          & 1.91           & 9.64          & 1.37 & \multicolumn{1}{c|}{{3.90}}    & 2.07          & 8.51          \\
Corrformer   & 1.99          & \multicolumn{1}{c|}{8.21}          & 2.09          & \multicolumn{1}{c|}{9.47}          & 1.38          & \multicolumn{1}{c|}{3.83}          & 66.7          & \multicolumn{1}{c|}{7832.3}          & 2.19           & 12.39         & \textbf{1.30} & \multicolumn{1}{c|}{3.89}          & {\ul 1.89}    & {\ul 7.71}    \\
iTransformer & {\ul 1.64}    & \multicolumn{1}{c|}{{\ul 5.94}}    & {\ul 1.67}    & \multicolumn{1}{c|}{{\ul 6.57}}    & 1.24          & \multicolumn{1}{c|}{3.31}          & 58.6          & \multicolumn{1}{c|}{6570.6}          & 1.47           & 5.53          & {\ul 1.33}          & \multicolumn{1}{c|}{3.87}          & 1.90          & \textbf{7.50} \\
RPMixer      & 1.77          & \multicolumn{1}{c|}{6.60}          & 1.83          & \multicolumn{1}{c|}{7.45}          & 1.28          & \multicolumn{1}{c|}{3.52}          & 60.3          & \multicolumn{1}{c|}{6607.1}          & 1.65           & 6.45          & 1.43          & \multicolumn{1}{c|}{4.02}          & 2.47          & 11.08         \\
PatchSTG     & 1.65          & \multicolumn{1}{c|}{{\ul 5.94}}    & 1.68          & \multicolumn{1}{c|}{6.58}          & {\ul 1.20}    & \multicolumn{1}{c|}{{\ul 3.15}}    & {\ul 57.2}    & \multicolumn{1}{c|}{{\ul 6254.8}}    & 1.41           & 4.92          & 1.36          & \multicolumn{1}{c|}{3.89}          & 2.21          & 9.59          \\ \midrule
S$^2$Transformer  & \textbf{1.47} & \multicolumn{1}{c|}{\textbf{4.99}} & \textbf{1.52} & \multicolumn{1}{c|}{\textbf{5.67}} & \textbf{1.17} & \multicolumn{1}{c|}{\textbf{3.09}} & \textbf{55.5} & \multicolumn{1}{c|}{\textbf{6135.7}} & {\ul 1.26}     & {\ul 4.08}    & \textbf{1.30} & \multicolumn{1}{c|}{\textbf{3.61}} & \textbf{1.87} & \textbf{7.50} \\ \bottomrule
\end{tabular}}
\end{small}
\end{center}
\end{table*}

Table~\ref{table:performance-comparison} presents the average forecast performance in 24 hours of all methods with an input length of 48 hours. Notably, following previous studies, the models on the WEATHER-5K dataset adopt a unified architecture to predict all variables at once, while those on the NCEI Global dataset are trained and tested separately for each variable. The best results are highlighted in \textbf{bold} and the second-best in \uline{underlined}. To ensure a fair comparison, all baselines are implemented with their official configurations. The experimental conclusions are as follows:

First, although current physics-based NWP models are regarded as the most accurate weather forecasting models, time series forecasting methods have demonstrated comparable performance in short-term forecasting tasks. Second, among TSF methods, the suboptimal performance of Informer, Autoformer, Pyraformer, and STID highlights the critical role of spatial correlation modeling. In contrast, spatial correlation modeling methods (MTGNN, iTransformer, and Corrformer) perform more effectively on the medium-scale NCEI Global dataset, underscoring the significance of this modeling paradigm. Our method outperforms all baselines, delivering up to a \textbf{5.7\%} performance improvement on the Wind variable of the NCEI Global dataset, attributed to its structured dynamic spatial correlation modeling. Furthermore, the limitations of the aforementioned models, such as the quadratic complexity of MTGNN and iTransformer, and the large parameter scale of Corrformer due to its complex structure, restrict their application to the large-scale WEATHER-5K dataset, further emphasizing the advantages of scalable spatial correlation modeling approaches. However, methods that simplify spatial receptive fields (PatchSTG) and mix channels in the spatial dimension (RPMixer) both suffer from information loss, caused by insensitivity to spatial structural information or spatial downsampling. In contrast, our method achieves state-of-the-art performance across most variables on the WEATHER-5K dataset, with an improvement of up to \textbf{16.8\%} on the Temperature variable. This superior performance stems from the structured spatial correlation constraints of our method: accurate perception and capture of local spatial correlations and restricted global spatial correlations, which preserve key information and filter out noise. We further provide the result and analysis of long term global station weather forecasting in the Appendix~\ref{Long term Forecasting Result}.
 
\subsection{Efficiency Study}
\label{Efficiency Study}

\begin{table*}[]
\caption{Global station weather forecasting efficiency comparison. BS: batch size used in model training. Mem: max memory used during training (in gigabytes). Time: total training time (in hours). The best results are highlighted in \textbf{bold}.}
\label{table:efficiency-analysis}
\begin{center}
\begin{small}
\begin{tabular}{l|ccc|ccc}
\toprule
Dataset          & \multicolumn{3}{c|}{WEATHER-5K} & \multicolumn{3}{c}{NCEI Global} \\ \midrule
Metric           & BS      & Mem        & Time     & BS      & Mem        & Time     \\ \midrule
MTGNN            & \textbf{64}      & 179.3      & 12.7     & \textbf{64}      & 64.3       & 0.6      \\
Corrformer       & 8       & 143.6      & 29.4     & 8       & 143.2      & 9.0      \\
iTransformer     & 16      & 151.2      & 7.5      & 24      & 159.5      & 0.3      \\ 
S$^2$Transformer & \textbf{64}      & \textbf{52.0}       & \textbf{5.0}      & \textbf{64}      & \textbf{43.3}       & \textbf{0.2}      \\ \bottomrule
\end{tabular}
\end{small}
\end{center}
\end{table*}

First, our proposed model consists of two stages: preprocessing and training. Verified by multiple experiments, the preprocessing time for both datasets is stably within 2 minutes. Compared with the training duration, the preprocessing time is negligible. To further evaluate model efficiency, we compare our method with existing spatial correlation modeling methods in three key metrics: \textbf{batch size}, \textbf{maximum memory usage during training}, and \textbf{total training time consumption}. The models are trained with multi-GPU parallel acceleration using \textit{torch.nn.DataParallel}. As shown in Table~\ref{table:efficiency-analysis}, the key observations are as follows.

Existing spatial correlation modeling models (MTGNN, iTransformer) exhibit rapid growth in computational and/or memory consumption as the number of nodes increases, primarily due to their quadratic complexity. Although Corrformer only models unidirectional spatial correlations, its complex Encoder-Decoder architecture (different from the aforementioned Encoder-Only models) leads to a large memory footprint, which further results in a smaller batch size and longer training time. In contrast, the proposed model in this paper confines fine-grained spatial correlation modeling within local subgraphs, enabling efficient long-distance information exchange via inter-subgraph spatial correlation modeling. This design not only filters out noise to improve performance but also effectively reduces memory consumption. Across the two datasets, the efficiency ranking of all models remains consistent: S$^2$Transformer > iTransformer > MTGNN > Corrformer. Particularly, compared with Corrformer (specifically designed for GSWF tasks), S$^2$Transformer reduces memory consumption by approximately \textbf{64}\% and improves inference speed by about \textbf{83}\% on the WEATHER-5K dataset; on the NCEI Global dataset, it achieves a memory reduction of around \textbf{70}\% and an inference speedup of roughly \textbf{98}\%.

\subsection{Ablation Study}
\label{Ablation Study}

\begin{table}[]
\caption{Ablation study on the WEATHER-5K dataset. The best results are highlighted in \textbf{bold}.} 
\label{table:ablation-study}
\begin{center}
\begin{small}
\resizebox{\linewidth}{!}{
\begin{tabular}{l|ll|ll|ll|ll|ll}
\toprule
Variable       & \multicolumn{2}{c|}{Temperature} & \multicolumn{2}{c|}{Dewpoint} & \multicolumn{2}{c|}{Wind Rate} & \multicolumn{2}{c|}{Wind Direc.}  & \multicolumn{2}{c}{Sea Level} \\ \midrule
Metric         & MAE             & MSE            & MAE           & MSE           & MAE            & MSE           & MAE            & MSE              & MAE           & MSE           \\ \midrule
w/o. Metis     & 1.64            & 5.99           & 1.71          & 6.83          & 1.25           & 3.40          & 59.25          & 6607.08          & 1.72          & 7.88          \\
w/o. SH        & 1.49            & 5.08           & 1.54          & 5.75          & 1.18           & \textbf{3.09} & 56.02          & 6206.36          & 1.30          & 4.37          \\
w/o. Intra-Att & 1.65            & 6.03           & 1.70          & 6.77          & 1.26           & 3.44          & 59.74          & 6639.95          & 1.53          & 6.02          \\
w/o. Inter-Att & 1.57            & 5.63           & 1.62          & 6.34          & 1.23           & 3.30          & 58.44          & 6493.64          & 1.52          & 6.30          \\
w/o. SA        & 1.48            & 5.03           & 1.54          & 5.72          & 1.18           & 3.06          & 55.75          & 6171.15          & \textbf{1.26} & 4.13          \\ \midrule
S$^2$Transformer  & \textbf{1.47}   & \textbf{4.99}  & \textbf{1.52} & \textbf{5.67} & \textbf{1.17}  & \textbf{3.09} & \textbf{55.50} & \textbf{6135.69} & \textbf{1.26} & \textbf{4.08} \\ \bottomrule
\end{tabular}}
\end{small}
\end{center}
\end{table}

We perform an ablation study on the WEATHER-5K dataset to validate the effectiveness of the proposed modules. Specifically, we consider the following variants of our proposed model:
(1) \textbf{w/o. Metis}: The METIS graph partitioning method is substituted with random partitioning. (2) \textbf{w/o. SH}: The spherical harmonic position encoding in the preprocessing stage is removed. (3) \textbf{w/o. Intra-Att}: The intra-subgraph attention is removed, and only the spatial correlation between subgraphs is captured. (4) \textbf{w/o. Inter-Att}: The inter-subgraph attention is removed, and only the local spatial correlation is captured. (5) \textbf{w/o. SA}: The spatial attention bias matrix in both inter-subgraph and intra-subgraph attention is removed. As shown in Table~\ref{table:ablation-study}, the key observations are as follows.

First, the performance drop of \textbf{w/o. Metis} confirms the importance of graph partitioning algorithms. METIS, a hierarchical partitioning algorithm, ensures subgraph balance with only the partition count as a parameter. Second, \textbf{w/o. Intra-Att} shows the most severe performance decline, indicating that local spatial correlation is critical for accurate global station weather forecasting, consistent with Tobler’s first law of geography. Meanwhile, \textbf{w/o. Inter-Att} exhibits the second-most significant drop, suggesting global spatial correlation also aids precise forecasting, aligning with Tobler’s second law of geography. Finally, the performance of \textbf{w/o. SH} and \textbf{w/o. SA} validates the effectiveness of the proposed spherical harmonic position encoding and spatial attention bias.

\subsection{Parameter Sensitivity Analysis}
\label{Parameter Sensitivity Analysis}

\begin{figure} 
\centering    
\subfigure[Blocks on Wind] {    
\includegraphics[width=0.35\columnwidth]{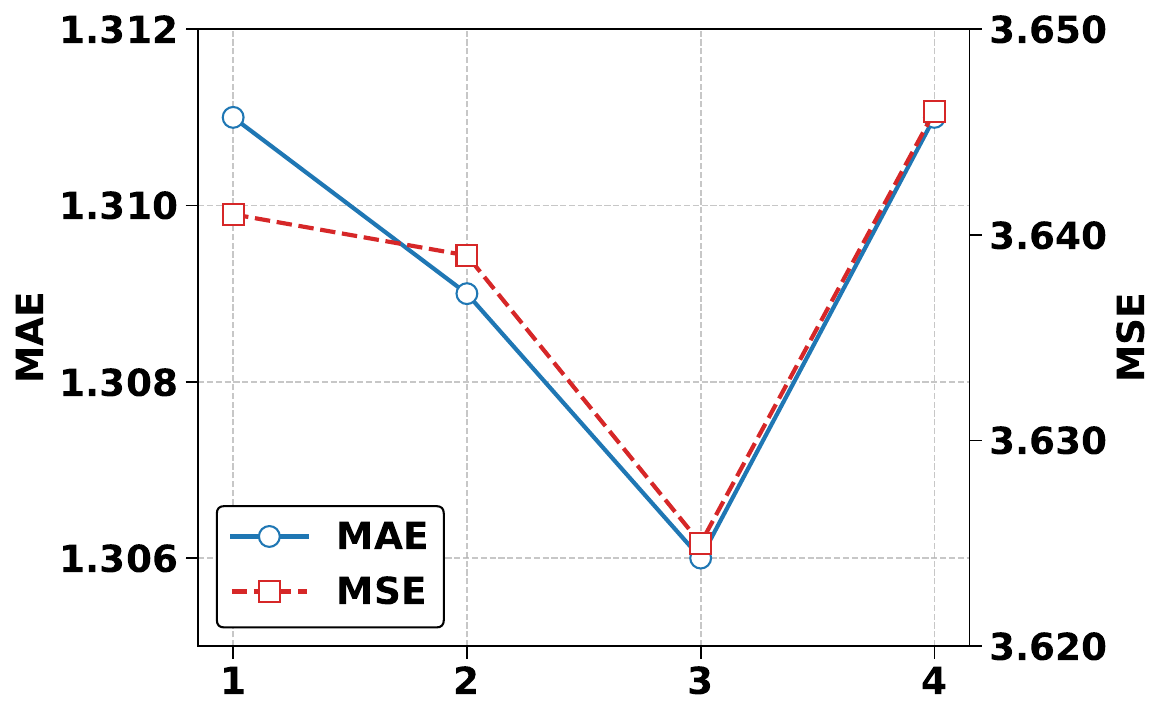}  
}     
\subfigure[Blocks on Temp] {     
\includegraphics[width=0.35\columnwidth]{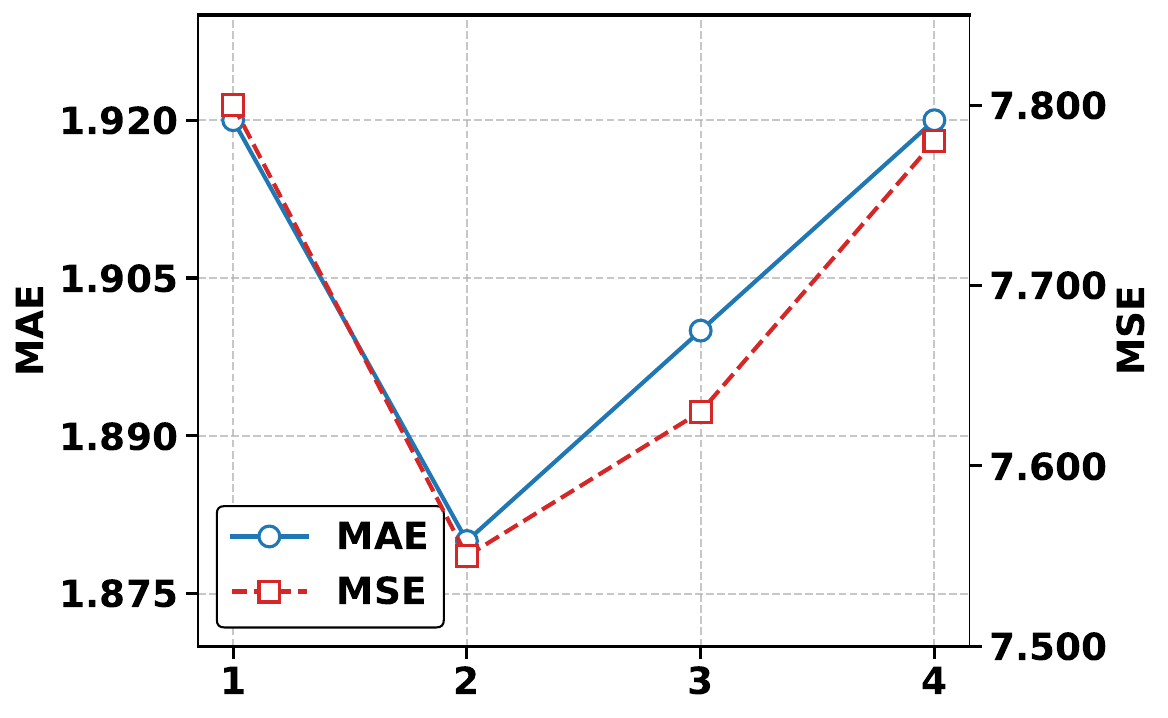}     
}    
\subfigure[Dimension on Wind] {     
\includegraphics[width=0.35\columnwidth]{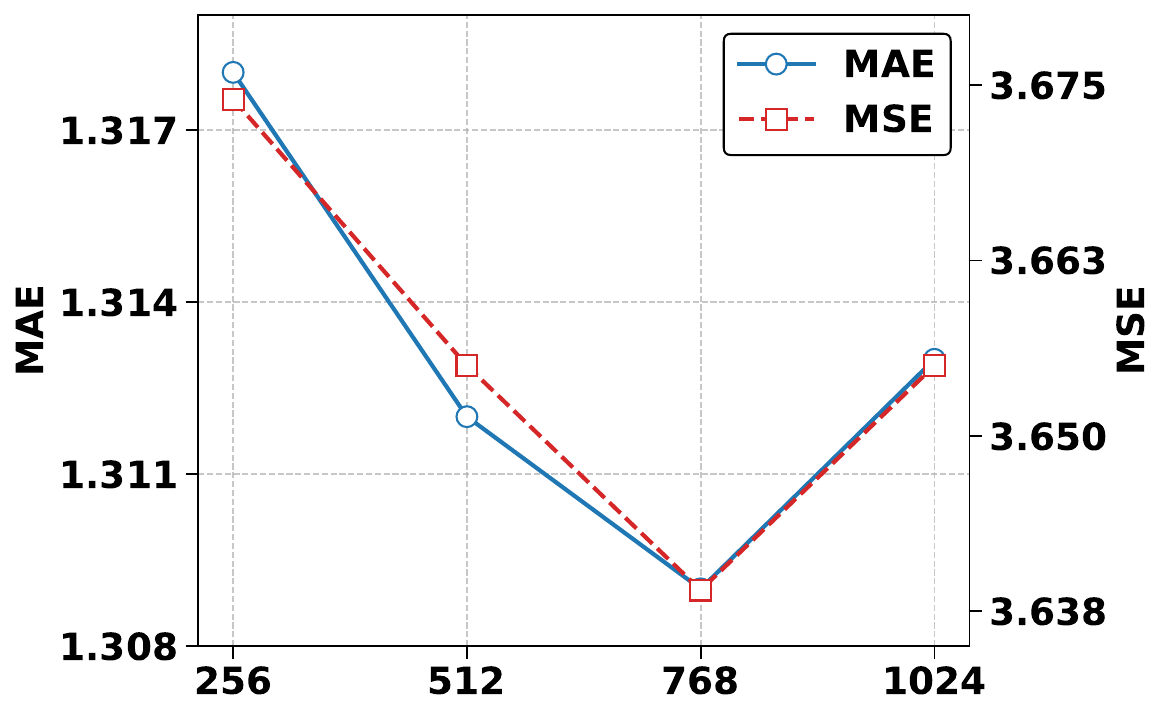}     
}    
\subfigure[Dimension on Temp] {     
\includegraphics[width=0.35\columnwidth]{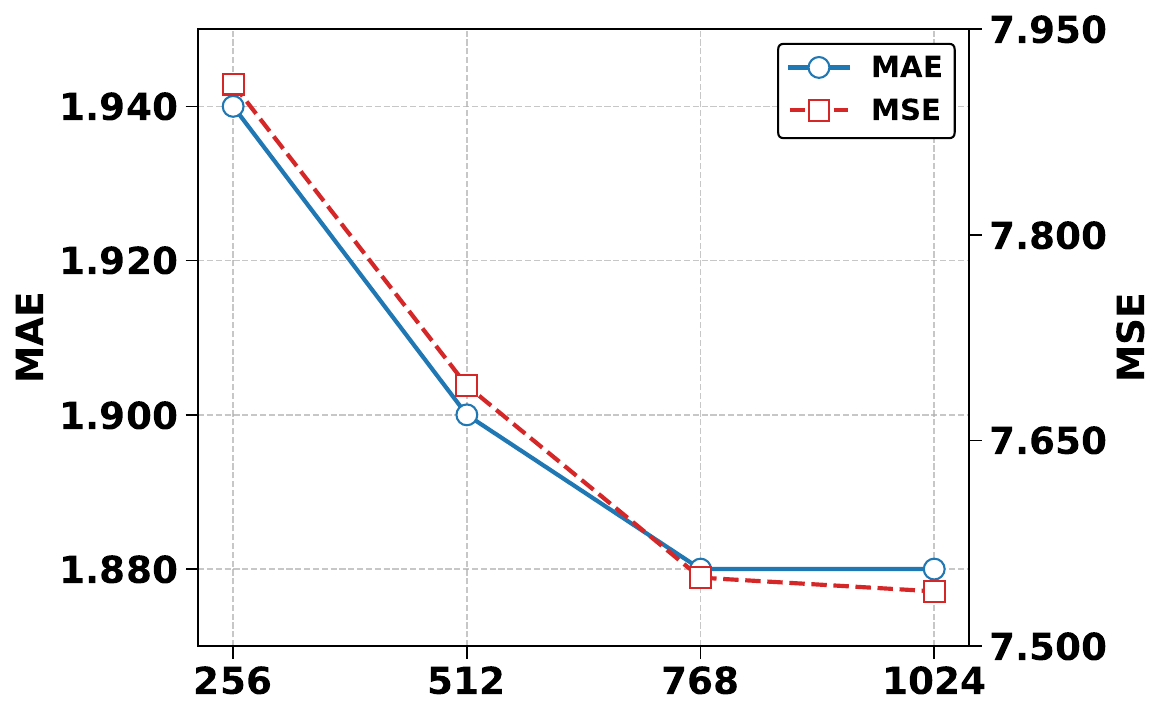}    
}    
\caption{Parameter sensitivity analysis}     
\label{fig:parameter-layer-hiddenchannel}     
\end{figure}

We evaluate the sensitivity of hyperparameters (including the number of blocks $L$ and embedding dimension $D$) on the NCEI Global dataset. Figure \ref{fig:parameter-layer-hiddenchannel} shows that model performance improves with increasing $L$, achieving the best results at \(L=2\) for the Global Temperature dataset and \(L=3\) for the Global Wind dataset. Further increasing $L$ leads to a degradation in performance. Thus, we select \(L=2\) for model efficiency. Similarly, optimal performance is achieved with an embedding dimension of $D=768$, with no further performance gain from increasing $D$. This indicates that a small model suffices to learn spatiotemporal knowledge in global station weather forecasting. The sensitivity analysis of look-back window length \(T\) and the number of subgraphs $P$ is presented in Appendix~\ref{More Parameters Sensitivity Analysis}. 

\subsection{Visualization}
\label{Visualization}

\begin{figure}
    \centering
    \includegraphics[width=0.58\textwidth]{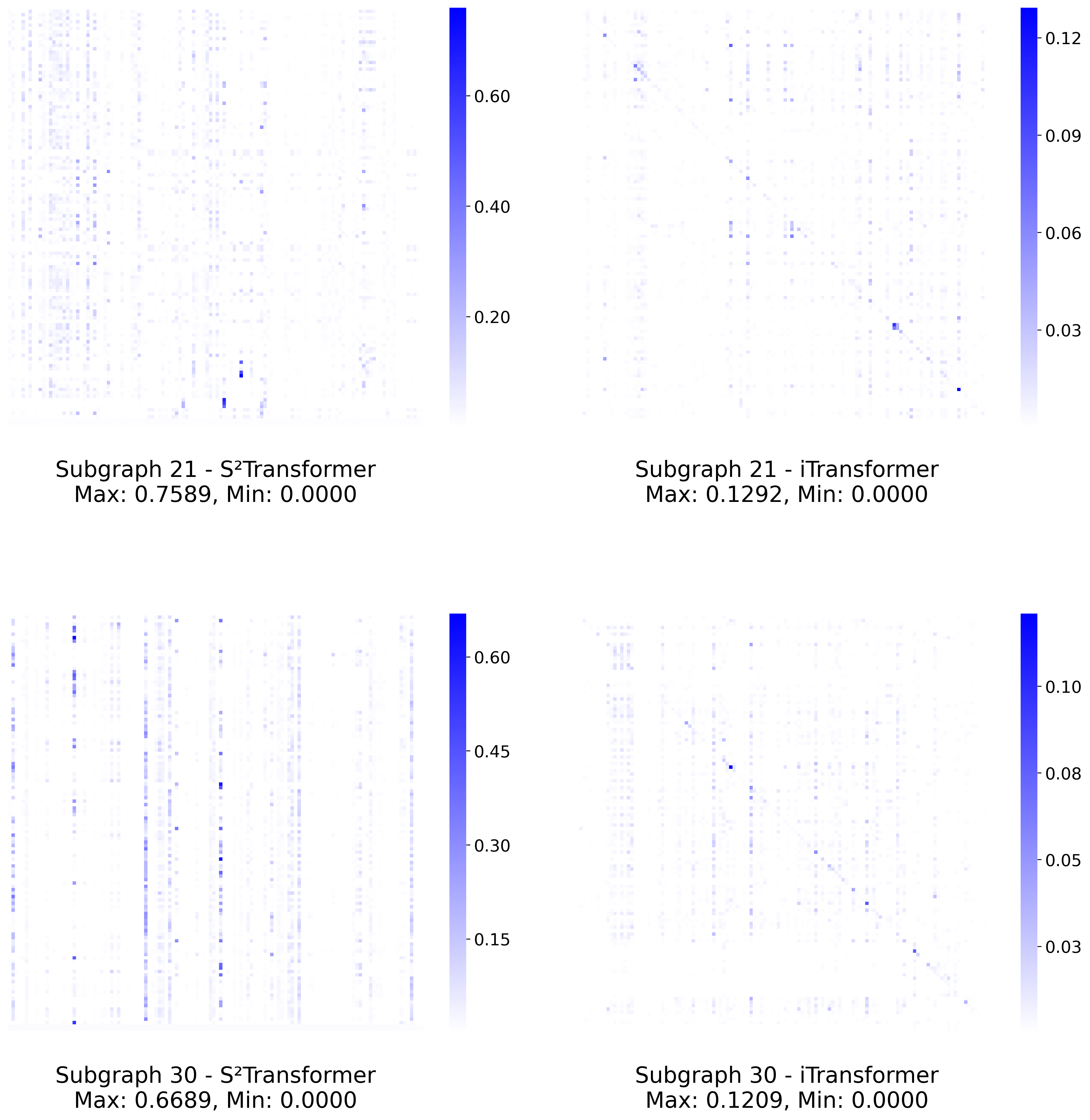}
    \caption{Visualization of intra-subgraph attention matrices for S$^2$Transformer and iTransformer.}
    \label{fig:intra-attention-visualization}
\end{figure} 

To gain deeper insights into our proposed model, we conducted supplementary visualization experiments, focusing on a comparative analysis of the intra-subgraph attention matrix of our model (S²Transformer) and the attention matrix of the iTransformer model. Specifically, we selected the intra-subgraph attention matrices corresponding to two subgraphs from the test data. For the purpose of comparison, we extracted the relevant subgraph attention matrix fragments from iTransformer’s global attention matrix, with the extraction process strictly guided by the node indices of the two aforementioned subgraphs.

As illustrated in Figure~\ref{fig:intra-attention-visualization}, owing to the global receptive field of iTransformer’s attention mechanism, the attention values in its corresponding subgraph regions are scattered and small (with a maximum of approximately 0.12). When the number of such scattered trivial values is large, they tend to introduce trivial noise and impair model performance. In contrast, our model proactively restricts the receptive field to confine attention within the subgraph—an approach that not only aligns with relevant geographical laws but also yields more concentrated and larger attention values (with a broader distribution of attention weights). This confirms that our model effectively avoids noise introduction during local modeling. The quantitative discrepancy between this structured spatial correlation and iTransformer’s unstructured attention serves as direct evidence of noise reduction, validating the rationality of our model design. We provide additional visualization analysis results in Appendix~\ref{More Visualization}.

\section{Conclusion and Limitation}
In this paper, we propose a novel Spatial Structured Attention Block that not only perceives spatial structure but also considers both spatial proximity and global correlation. Specifically, we partition the spatial graph into a set of subgraphs and utilize the Intra-subgraph and Inter-subgraph Attention to learn local and global spatial correlation. Building on the proposed block, we develop a multiscale model S$^2$Transformer by progressively increasing the subgraph scales. The resulting model is scalable and can produce structured spatial correlation. Performance comparison and efficiency analysis validate the superiority of our method in medium and large-scale global station weather forecasting scenarios. Ablation studies confirm the effectiveness of the model designs, and we further explore the hyperparameters in model construction.

\textbf{Limitation}. However, we noticed that our model still lags behind the numerical weather prediction model for longer lead times. This is mainly because an increase in forecast duration amplifies the non-linear dynamics of the atmospheric system, and pure data-driven methods struggle to fully capture their evolutionary laws. In the future, we would like to explore the nascent regime of data-driven and physics-informed paradigms to enhance the model's ability to predict long-term weather processes.

\subsubsection*{Broader Impact Statement}
In this paper, we propose a novel scalable spatiotemporal series forecasting model that captures structured spatial correlation guided by Tobler's laws of geography to enhance global station weather forecasting while maintaining low running costs. Our research aims to make a positive contribution to the relevant community while ensuring no negative social impact.



\bibliography{main}
\bibliographystyle{tmlr}

\appendix
\section{Scalable Spatiotemporal Series Forecasting}
\label{Scalable Spatiotemporal Series Forecasting}

Several researchers have developed scalable spatiotemporal forecasting methods to accommodate larger datasets. Model-agnostic approaches use graph partitioning to decompose large spatial graphs into smaller subgraphs, with experiments conducted via independent training \citep{mallick2020graph} or continual learning \citep{wangmake}. In contrast, existing scalable models fall into four paradigms: precomputing spatial correlation, linearizing spatial correlation computation, mixing channels across spatial dimensions, and simplifying nodes’ spatial receptive fields. SGP \citep{cini2023scalable} and SimST \citep{liu2024reinventing} precompute graph convolutions and decouple spatial correlation modeling from training, but their fixed input-space representations may reduce effectiveness. BigST \citep{han2024bigst} and Sumba \citep{chenstructured} adopt linearized spatial convolutions to lower complexity, yet low-rank approximations prevent them from capturing structured spatial correlation. The channel mixing approach \citep{yeh2024rpmixer,han2024softs} enhances scalability by aggregating and routing messages across dimensions, avoiding quadratic complexity but suffering from information dilution during aggregation, leading to suboptimal practical performance. Methods simplifying spatial receptive fields, such as SAGDFN \citep{jiang2024sagdfn}, use significant neighbor sampling to model spatial correlation but fail to preserve local structural information. Similarly, PatchSTG \citep{DBLP:conf/kdd/0001L0S0LJ025} partitions traffic nodes into non-overlapping KDTree-based patches, using depth attention for local correlation and breadth attention across same-index patches for global aggregation. However, lacking inherent integration of spatial prior knowledge (e.g., topological connections, geographic proximity), it overly relies on training patterns and struggles to capture critical geographic dependencies.

\section{Spherical Harmonics}
\label{Spherical Harmonics}

Spherical harmonics have been widely used in Earth science \citep{klosko1982spherical,pail2011first,thebault2021spherical}. Any function \( f(\lambda, \phi) \) defined on a sphere can be expressed as a weighted sum of orthogonal spherical harmonic basis functions \( {Y_l^m} \) with increasing frequency, characterized by degrees \( l \) and orders \( m \):  
\begin{equation}
    f(\lambda, \phi) = \sum_{l=0}^{\infty} \sum_{m=-l}^{l} w_l^m {Y_l^m} (\lambda, \phi) 
\end{equation}
Here, \( w_l^m \) are the weights. Each spherical harmonic \( {Y_l^m} \) is defined as:  
\begin{equation}
{Y_l^m} (\lambda, \phi) = \sqrt{\frac{2l+1}{4\pi} \cdot \frac{(l-|m|)!}{(l+|m|)!}} \, {P_l^{m}} (\cos\lambda) \, e^{im\phi} 
\label{eq:ylm}
\end{equation}
where \( {P_l^m(x)} \) denotes the associated Legendre polynomials, given by:  
\begin{equation}
{P_l^m(x)} = (-1)^m (1-x^2)^{\frac{m}{2}} \frac{d^{m}}{dx^{m}} P_l(x)
 \end{equation}
These involve derivatives of Legendre polynomials \( P_l(x) \), which are defined as: 
\begin{equation}
P_l(x) = \frac{1}{2^l l!} \frac{d^l}{dx^l} (x^2 - 1)^l
\end{equation}
For associated Legendre polynomials of negative order (\( m < 0 \)), the symmetry relation can be used.
\begin{equation}
P_l^{-m}(x) = (-1)^m \frac{(l - m)!}{(l + m)!} P_l^m(x)
\end{equation}
In practice, we use the real form of Eq.~\ref{eq:ylm}:  
\begin{equation}
Y_l^m(\lambda, \phi) = 
\begin{cases} 
(-1)^m \sqrt{2} \bar{P}_l^{|m|}(\cos\lambda) \sin(|m|\phi), & \text{if } m < 0, \\
\bar{P}_l^m(\cos\lambda), & \text{if } m = 0, \\
(-1)^m \sqrt{2} \bar{P}_l^m(\cos\lambda) \cos(m\phi), & \text{if } m > 0,
\end{cases} 
\end{equation}
where \( \bar{P}_l^m(\cos\lambda) \) denotes the normalized associated Legendre polynomial, defined as:  
\begin{equation}
\bar{P}_l^m(\cos\lambda) = \sqrt{\frac{2l+1}{4\pi} \cdot \frac{(l-|m|)!}{(l+|m|)!}} P_l^m(\cos\lambda)
\end{equation}

\section{Implementation Details}
\label{Implementation Details}

\subsection{Datasets}

\begin{figure} 
\centering    
\subfigure[WEATHER-5K] {    
\includegraphics[width=0.45\columnwidth]{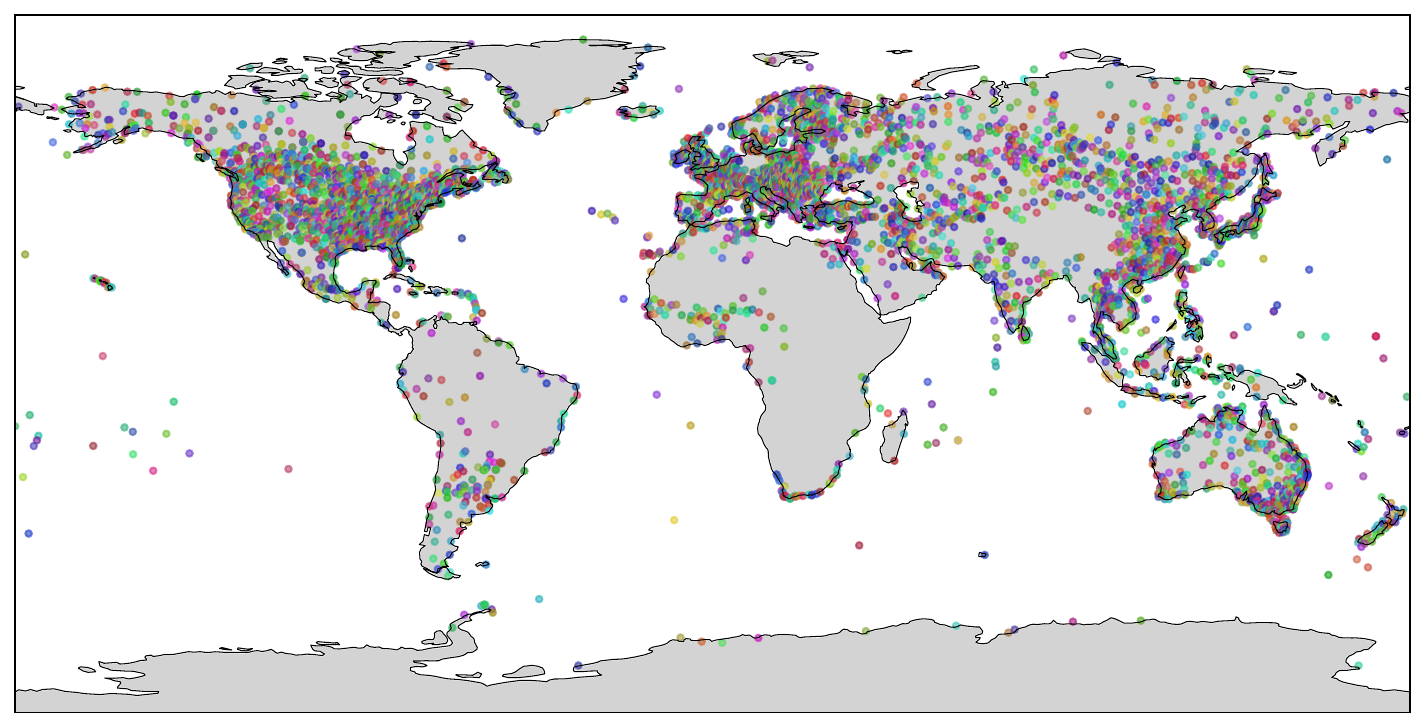}  
}     
\subfigure[NCEI Global] {     
\includegraphics[width=0.45\columnwidth]{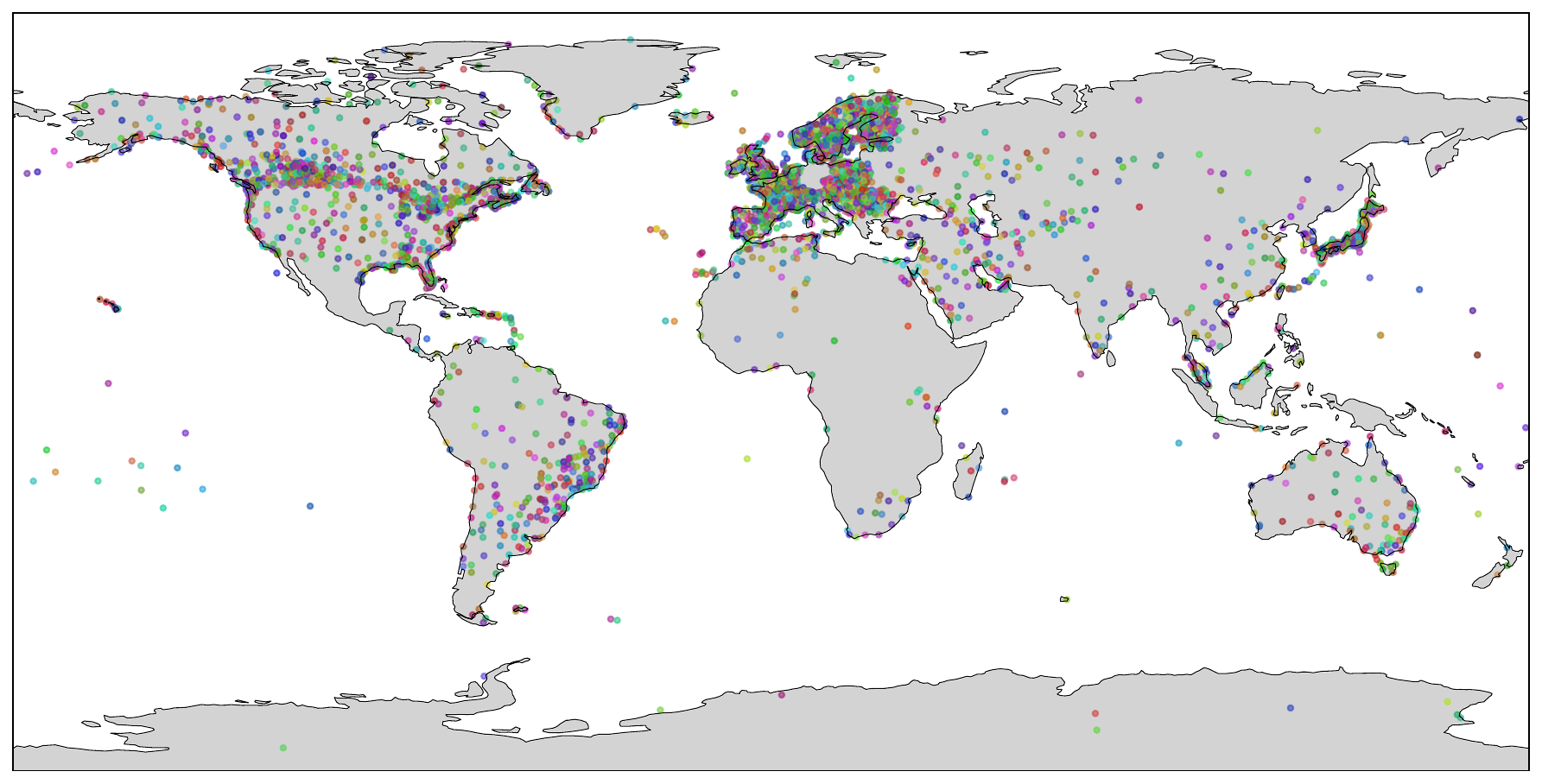}     
}    
\caption{Global meteorological station distribution}     
\label{fig:station-distribution}     
\end{figure}

All datasets used in our study are open-source or freely available for research purposes.

\textbf{WEATHER-5K}. The WEATHER-5K dataset, derived from global near-surface in-situ observations via the public Integrated Surface Database (ISD), includes data from 5,672 high-quality weather stations (2014-2023) covering variables like wind speed, direction, and temperature. It undergoes rigorous quality control (data interpretation, temporal alignment, completeness filtering, outlier detection), with remaining missing data interpolated using ERA5, followed by standardization and extreme value percentile calculation. Featuring uneven station distribution, it better aligns with actual observations than simulated ERA5 data, addressing limitations of small-scale existing time-series meteorological datasets for relevant research. The dataset can be accessed at \url{https://github.com/taohan10200/WEATHER-5K}.

\textbf{NCEI Global}. The NCEI Global dataset, sourced from the National Centers for Environmental Information, comprises hourly averaged wind speed and temperature records from 3,850 stations across the globe, covering the period from January 1, 2019, to December 31, 2020. It is divided into two subsets: "global wind" and "global temp". The dataset is publicly accessible via the National Oceanic and Atmospheric Administration (NOAA) at \url{https://www.ncei.noaa.gov/data/global-hourly/access}. We utilize the available processed versions in this study, which are available in the Corrformer GitHub repository: \url{https://github.com/thuml/Corrformer}.

We visualize the station distributions of both datasets in~\cref{fig:station-distribution}, which shows that the stations effectively cover diverse weather patterns across varying geographical scales and station densities. Following previous research, we chronologically split the WEATHER-5K dataset into training, validation, and test sets at a ratio of 0.8/0.1/0.1, and the NCEI Global dataset at 0.7/0.1/0.2. Input data were normalized using the Z-score for model training.

\subsection{Baselines}
We compare the proposed approach with physics-based NWP models and the following advanced time series forecasting baselines:
\begin{itemize}
    \item Informer: It utilizes ProbSparse self-attention to reduce computational complexity, enabling efficient long-sequence forecasting by focusing on dominant temporal patterns.
    \item Autoformer: It employs a decomposition-attention architecture to model trend and seasonal components, leveraging an auto-correlation mechanism to capture long-range dependencies without explicit alignment.
    \item Pyraformer: It uses a hierarchical pyramid graph to capture multi-scale temporal dependencies with linear complexity, enabling efficient long-range forecasting.
    \item STID: It combines spatial and temporal identity embeddings with multi-layer perceptrons to address sample indistinguishability in spatiotemporal dimensions.
    \item MTGNN: It builds an adaptive static global directed graph using learnable node embedding and aggregates information along spatial dimensions through mix-hop propagation.
    \item Corrformer: It integrates multi-correlation mechanisms (spatial cross-correlation and temporal auto-correlation) in a learnable tree structure to model complex spatiotemporal dependencies for large-scale global station weather forecasting.
    \item iTransformer: It embeds the whole time series into a spatial token and captures dynamic global spatial correlation using the self-attention mechanism.
    \item RPMixer: It employs MLPs to model temporal dependency and integrates random projection layers to capture spatial correlation.
    \item PatchSTG: It uses irregular spatial patching via KDTree and dual attention (depth and breadth) to capture local and global spatial correlations in large-scale spatiotemporal networks.
\end{itemize}
We obtain the code of baselines directly from their corresponding GitHub repositories. For the model- and training-related configurations, we follow the recommended settings provided in their code.

\subsection{Hyperparameters Setting}

To better reproduce our model, we summarize all the default hyperparameters as follows. The dimensions of the input embedding and hidden embedding dimension \(D\) are set to 768. The number of blocks \(L\) is set to 2. The initial number of subgraphs $P$ is set based on hyperparameter tuning results. Specifically, 64 for the WATHER-5K and NCEI Global Wind dataset, and 16 for the NCEI Global Temp dataset. In the calculation of spherical harmonic basis functions, we set the maximum order $l$ of Legendre polynomials to 3. The epsilon \(\epsilon\) is set by following the suggestion of DCRNN \cite{li2017diffusion}. The source code of our model will be available soon.

\section{Long term Forecasting Result}
\label{Long term Forecasting Result}

\begin{table*}[]
\renewcommand{\arraystretch}{1.5}
\caption{Long term global station weather forecasting performance comparison. The best results are highlighted in \textbf{bold}, while the second-best results are \uline{underlined}.} 
\label{table:long-term-performance-comparison}
\begin{center}
\begin{small}
\resizebox{\linewidth}{!}{
\begin{tabular}{l|l|ll|ll|ll|ll|ll}
\toprule
\multirow{2}{*}{Baselines}        & \multirow{2}{*}{\begin{tabular}[c]{@{}l@{}}Lead\\ Time\end{tabular}} & \multicolumn{2}{l|}{Temperature} & \multicolumn{2}{l|}{Dewpoint}  & \multicolumn{2}{l|}{Wind Rate} & \multicolumn{2}{l|}{Wind Direc.} & \multicolumn{2}{l}{Sea Level} \\ \cline{3-12} 
                                  &                                                                         & MAE            & MSE             & MAE           & MSE            & MAE            & MSE           & MAE            & MSE             & MAE           & MSE           \\ \midrule
\multirow{4}{*}{NWP Model}       & 24                                                                      & 1.76           & 7.39            & 1.85          & 7.94           & 1.48           & 4.53          & 63.8           & 7158.3          & \textbf{0.86} & \textbf{2.68} \\
                                  & 72                                                                      & \textbf{1.87}  & \textbf{8.01}   & \textbf{1.94} & \textbf{8.48}  & {\ul 1.52}     & {\ul 4.76}    & 72.4           & {\ul 8215.6}    & \textbf{1.06} & \textbf{3.31} \\
                                  & 120                                                                     & \textbf{1.99}  & \textbf{8.79}   & \textbf{2.14} & \textbf{10.87} & {\ul 1.58}     & \textbf{5.11} & 75.4           & 8647.7          & \textbf{1.38} & \textbf{5.15} \\
                                  & 168                                                                     & \textbf{2.15}  & \textbf{10.06}  & \textbf{2.32} & \textbf{12.56} & 1.66           & 5.59          & 78.3           & 8945.7          & \textbf{1.87} & \textbf{9.52} \\ \midrule
\multirow{4}{*}{Pyraformer}       & 24                                                                      & {\ul 1.75}     & {\ul 6.92}      & {\ul 1.83}    & {\ul 7.88}     & {\ul 1.30}     & {\ul 3.58}    & {\ul 61.8}     & {\ul 6930.2}    & 1.90          & 9.72          \\
                                  & 72                                                                      & 2.47           & 13.03           & 2.67          & 15.39          & 1.52           & 4.97          & {\ul 72.0}     & 8222.4          & 3.76          & 33.67         \\
                                  & 120                                                                     & 2.77           & 16.04           & 3.00          & 18.95          & 1.59           & 5.37          & {\ul 75.1}     & {\ul 8610.7}    & 4.43          & {\ul 43.91}   \\
                                  & 168                                                                     & 2.95           & 17.95           & 3.20          & 21.06          & {\ul 1.61}     & {\ul 5.56}    & {\ul 76.4}     & \textbf{8773.5} & {\ul 4.77}    & {\ul 49.97}   \\ \midrule
\multirow{4}{*}{S$^2$Transformer} & 24                                                                      & \textbf{1.47}  & \textbf{4.99}   & \textbf{1.52} & \textbf{5.67}  & \textbf{1.17}  & \textbf{3.09} & \textbf{55.5}  & \textbf{6135.7} & {\ul 1.26}    & {\ul 4.08}    \\
                                  & 72                                                                      & {\ul 2.20}     & {\ul 10.32}     & {\ul 2.36}    & {\ul 12.33}    & \textbf{1.46}  & \textbf{4.62} & \textbf{68.9}  & \textbf{7934.7} & {\ul 3.37}    & {\ul 27.54}   \\
                                  & 120                                                                     & {\ul 2.65}     & {\ul 14.54}     & {\ul 2.87}    & {\ul 17.48}    & \textbf{1.54}  & {\ul 5.13}    & \textbf{73.7}  & \textbf{8601.2} & {\ul 4.42}    & 44.56         \\
                                  & 168                                                                     & {\ul 2.87}     & {\ul 16.88}     & {\ul 3.12}    & {\ul 20.20}    & \textbf{1.59}  & \textbf{5.42} & \textbf{75.7}  & {\ul 8882.5}    & 4.92          & 53.55         \\ \bottomrule
\end{tabular}}
\end{small}
\end{center}
\end{table*}

As shown in Table~\ref{table:long-term-performance-comparison}, with the increase of forecasting lead time, the error of time series forecasting methods gradually increases: except for wind speed and wind direction, their performance on almost all variables is inferior to that of physics-based numerical weather prediction models. In contrast, NWP models produce more stable predictions—partly because they are typically trained on a larger scale and more abundant data, allowing them to deliver robust global atmospheric forecasts; partly because an increase in forecast duration amplifies the nonlinear dynamics of the atmospheric system, which pure data-driven methods struggle to fully capture. In the future, we intend to explore the emerging paradigm that integrates data-driven and physics-informed approaches to enhance the model’s capability of predicting long-term weather processes.
 
\section{More Parameters Sensitivity Analysis}
\label{More Parameters Sensitivity Analysis}

\begin{figure} 
\centering    
\subfigure[Lookback Length on Wind] {    
\includegraphics[width=0.35\columnwidth]{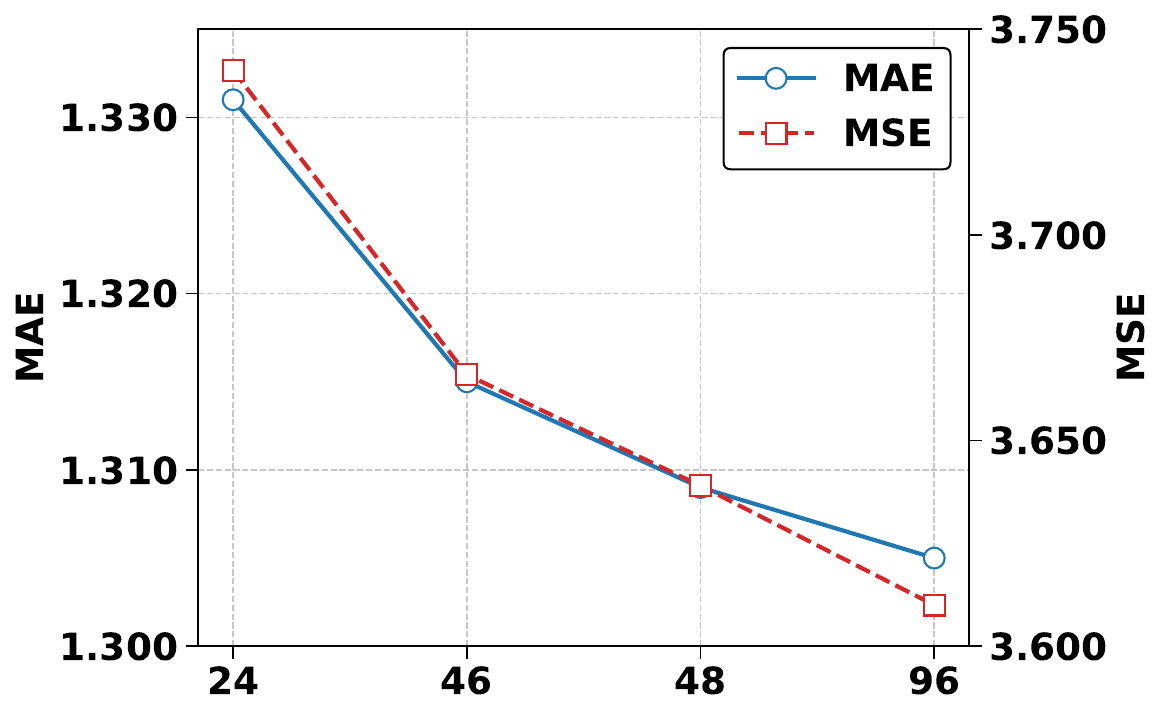}  
}     
\subfigure[Lookback Length on Temp] {     
\includegraphics[width=0.35\columnwidth]{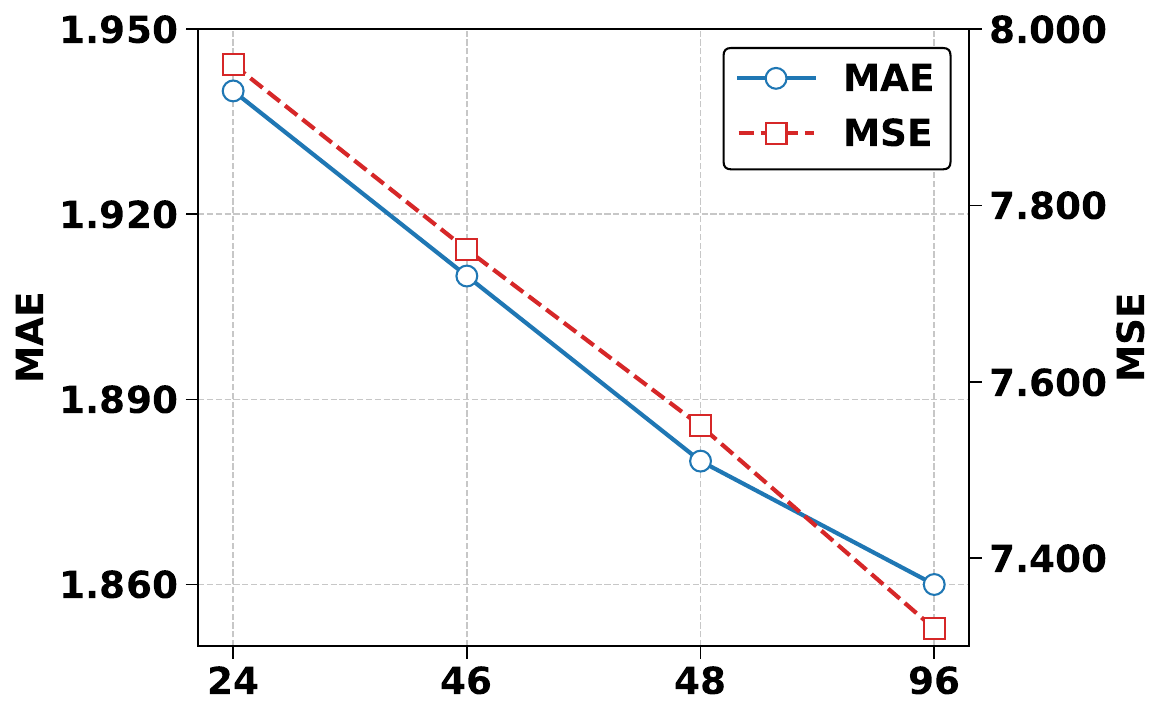}     
}    
\subfigure[Subgraph Number on Wind] {     
\includegraphics[width=0.35\columnwidth]{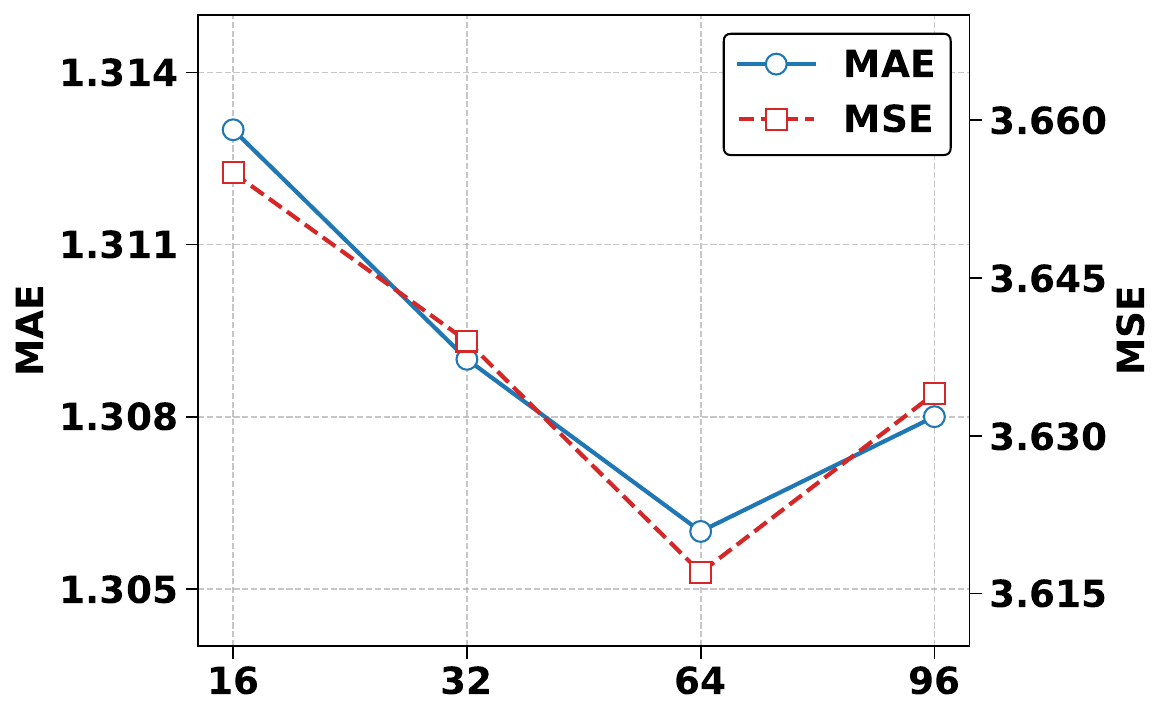}     
}    
\subfigure[Subgraph Number on Temp] {     
\includegraphics[width=0.35\columnwidth]{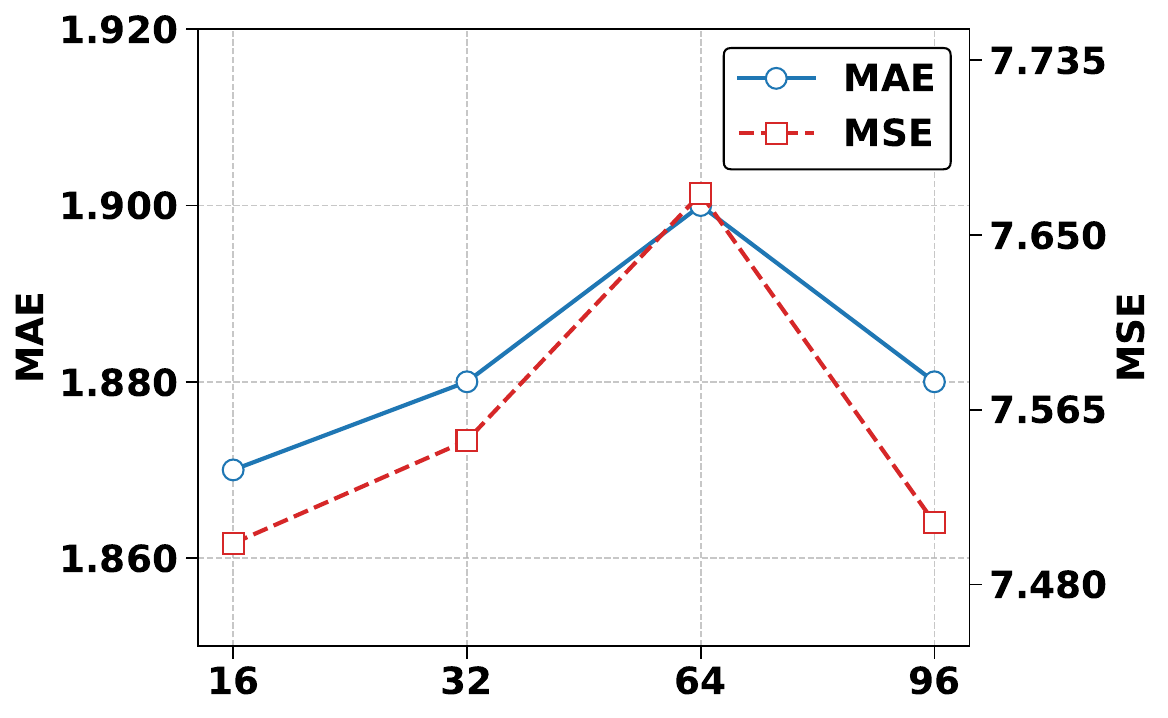}    
}    
\caption{More Parameter sensitivity analysis}     
\label{fig:parameter-lookback-subgraph}     
\end{figure}

We evaluate the impact of lookback window length \(T\) and initial subgraph number \(P\) on model performance on the NCEI Global dataset. As shown in Figure~\ref{fig:parameter-lookback-subgraph}, increasing \(T\) improves performance for both datasets. However, longer input sequences cause rapid surges in computational and memory costs for spatiotemporal GNNs, leading most existing models to rely on short-term historical windows, severely limiting their performance \citep{han2024bigst}. In contrast, our model achieves spatial scalability while accommodating larger lookback windows to boost performance. Next, we evaluate the impact of the initial subgraph number \(P\) (ranging from 16, 32, 64 to 96). A larger \(P\) means fewer nodes per subgraph, enabling more precise modeling of local spatial correlations; conversely, a smaller \(P\) increases nodes per subgraph, expanding the local spatial scope but raising memory and computational costs. Experimental results show that despite the same number of stations, the performance trends and optimal \(P\) values vary across datasets with different variables. This is attributed to complex interactions between local and global influences, indicating limitations in treating \(P\) merely as a hyperparameter without adjusting it to balance local and global effects. Exploring adaptive selection of \(P\) will be part of our future work.

\section{More Visualization}
\label{More Visualization}

\begin{figure}
    \centering
    \includegraphics[width=0.7\textwidth]{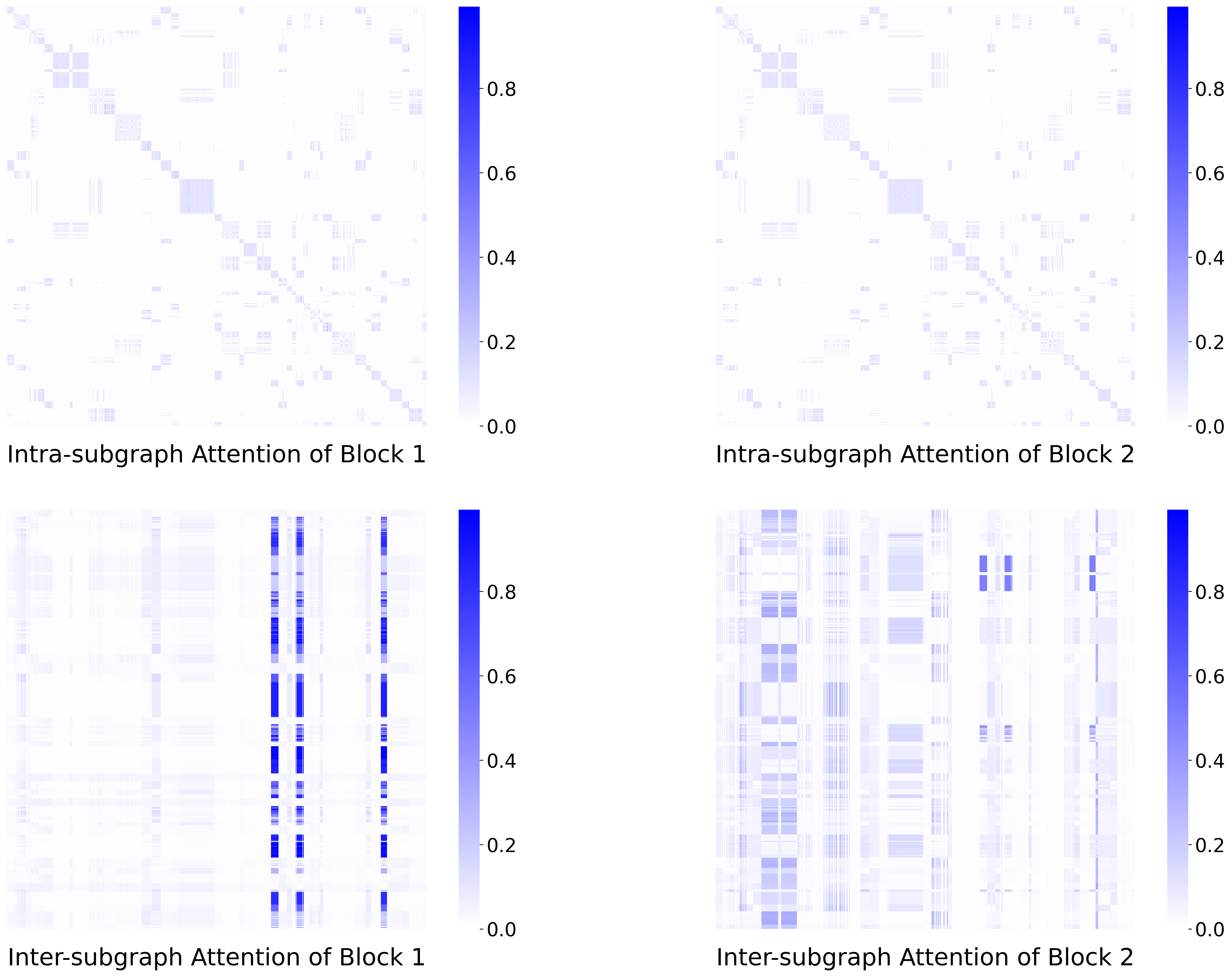}
    \caption{Visualization of intra-subgraph and inter-subgraph attention matrices for S$^2$Transformer.}
    \label{fig:full-model-attention-matrix}
\end{figure} 

\textbf{Attention matrix.} We further visualize the intra-subgraph and inter-subgraph attention matrices of S$^2$Transformer. As shown in Figure \ref{fig:full-model-attention-matrix}, the two-layer attention mechanism effectively separates local and global information (i.e., structured spatial correlation modeling): in local modeling, the model filters out global noise (the attention matrix has non-zero values only locally, showing a sparse high-rank pattern); while global modeling is achieved through subgraph aggregation and learning attention between subgraphs (the attention matrix exhibits high weights and a sparse low-rank pattern). We propose that this low-rank pattern can be interpreted as the discovery of key hubs in large-scale spatiotemporal networks.

\begin{figure} 
\centering    
\subfigure[iTransformer] {     
\includegraphics[width=0.7\columnwidth]{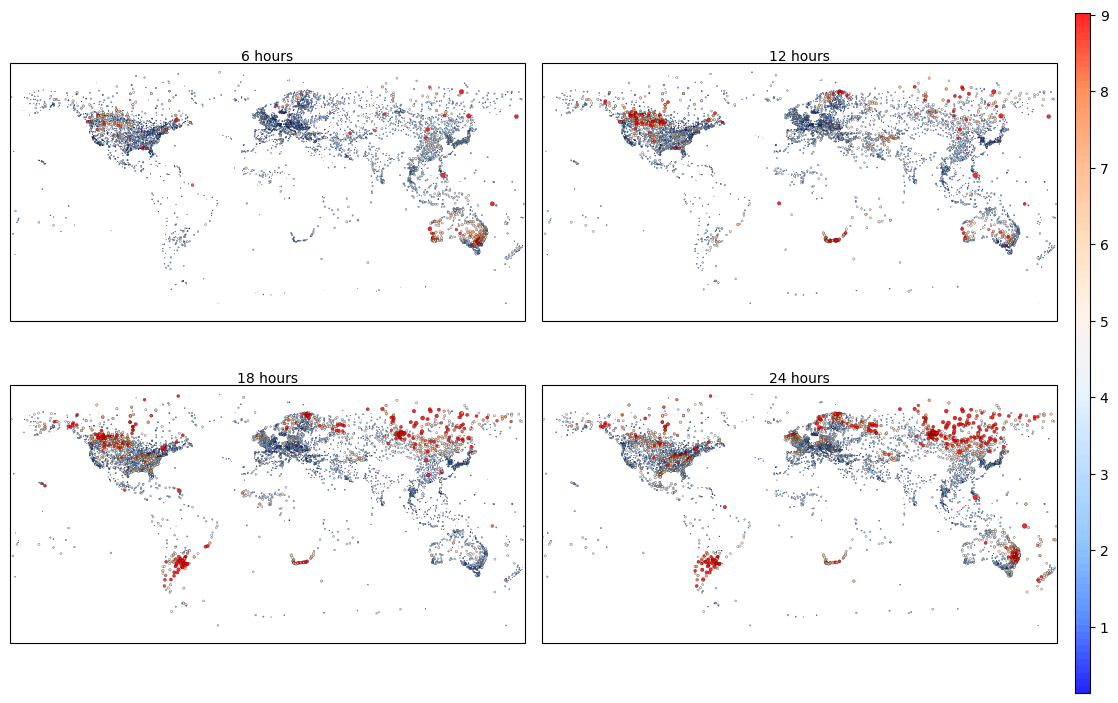}    
}    
\subfigure[S$^2$Transformer] {     
\includegraphics[width=0.7\columnwidth]{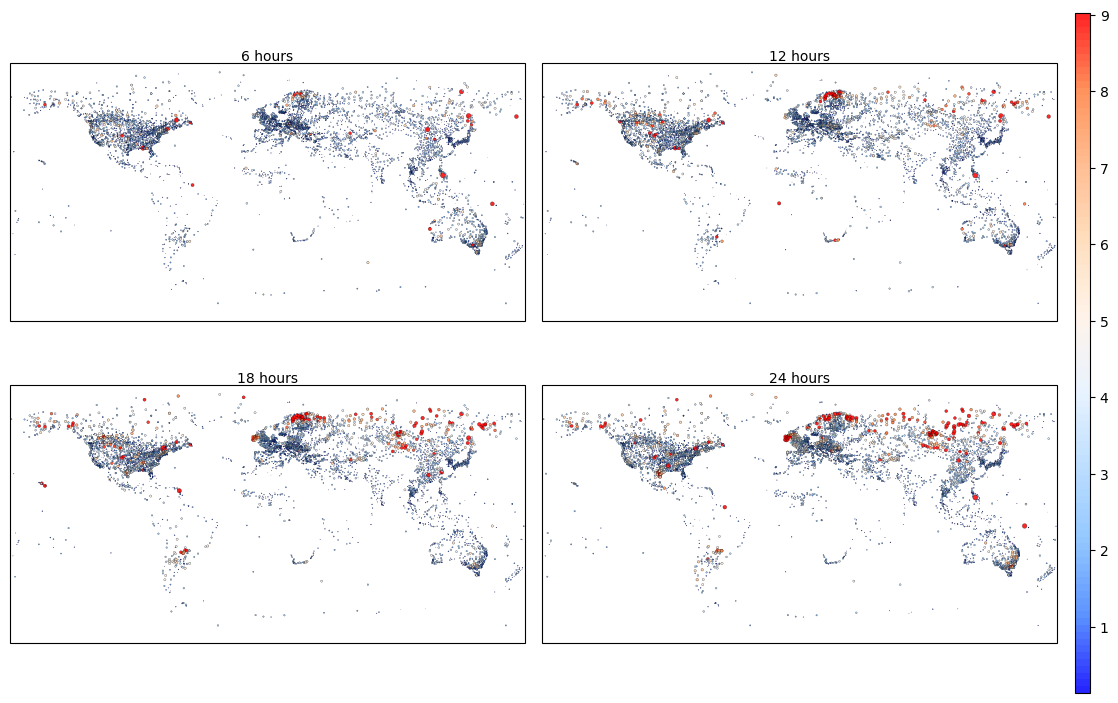}   
}     
\caption{Comparison of global station forecasting errors. }     
\label{fig:weather-5k-global-visualization}     
\end{figure}

\textbf{Global Station Forecasting result.} As shown in Figure \ref{fig:weather-5k-global-visualization}, we plot the prediction errors of different models for the temperature variable in the WEATHER-5K Dataset from a global perspective (brighter colors indicate larger prediction errors at the corresponding stations). It can be observed that all models show an upward trend in prediction errors over time; among them, S$^2$Transformer achieves higher accuracy in temperature prediction for high-latitude stations, while baseline models exhibit significantly larger prediction errors at these stations. This advantage originates from two designs of the proposed model: first, spherical harmonic positional encoding endows it with a clear definition worldwide (including polar regions), enabling better distinction of meteorological stations distributed globally; second, the attention mechanism with progressively expanded receptive fields conforms to physical diffusion laws.

\begin{figure} 
\centering    
\subfigure[STID] {     
\includegraphics[width=0.4\columnwidth]{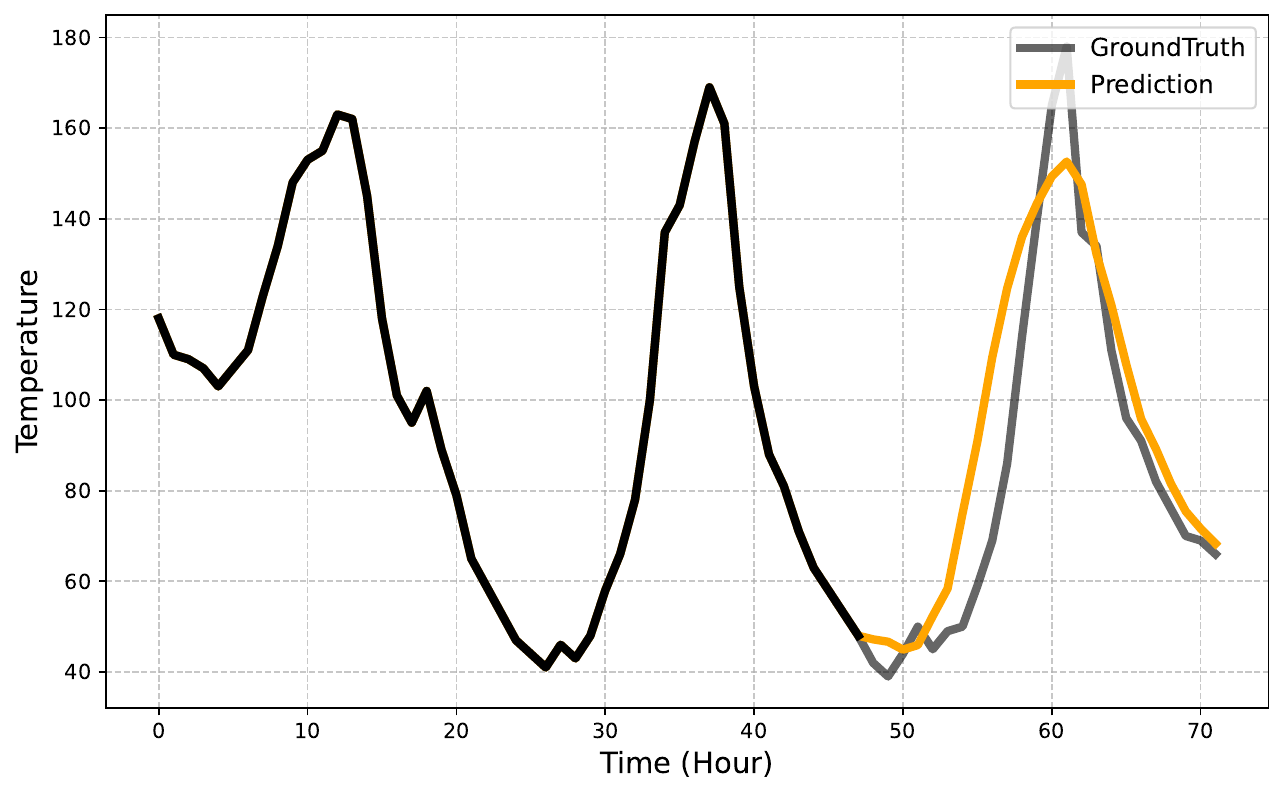}     
}    
\subfigure[RPMixer] {     
\includegraphics[width=0.4\columnwidth]{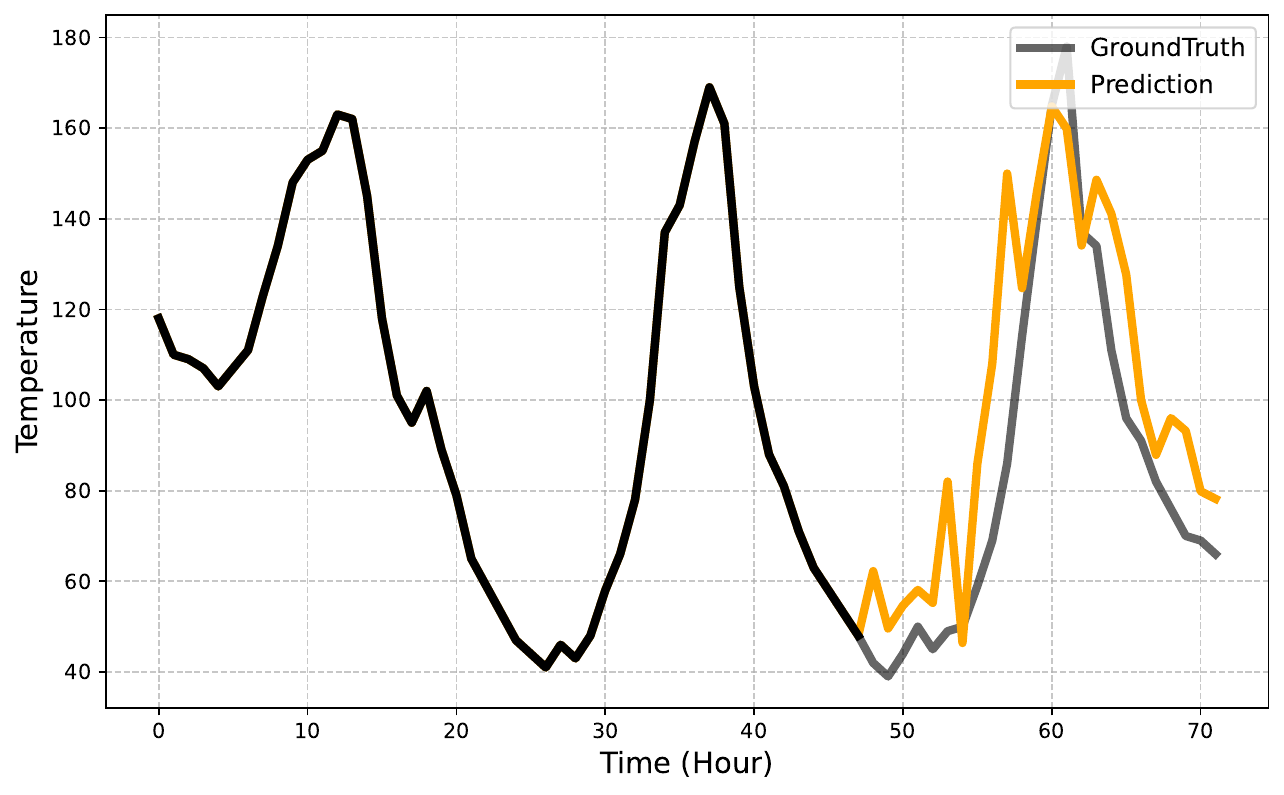}     
}    
\subfigure[PatchSTG] {     
\includegraphics[width=0.4\columnwidth]{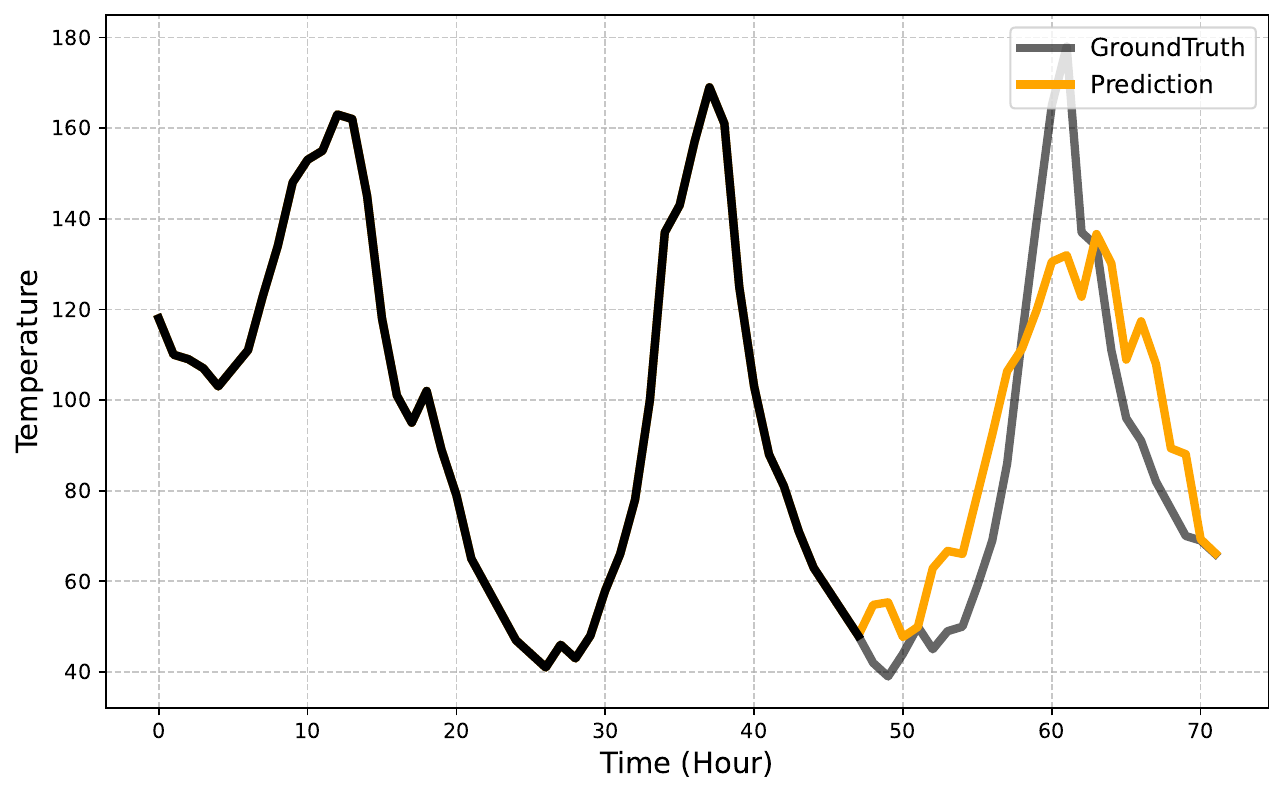}     
}    
\subfigure[S$^2$Transformer] {     
\includegraphics[width=0.4\columnwidth]{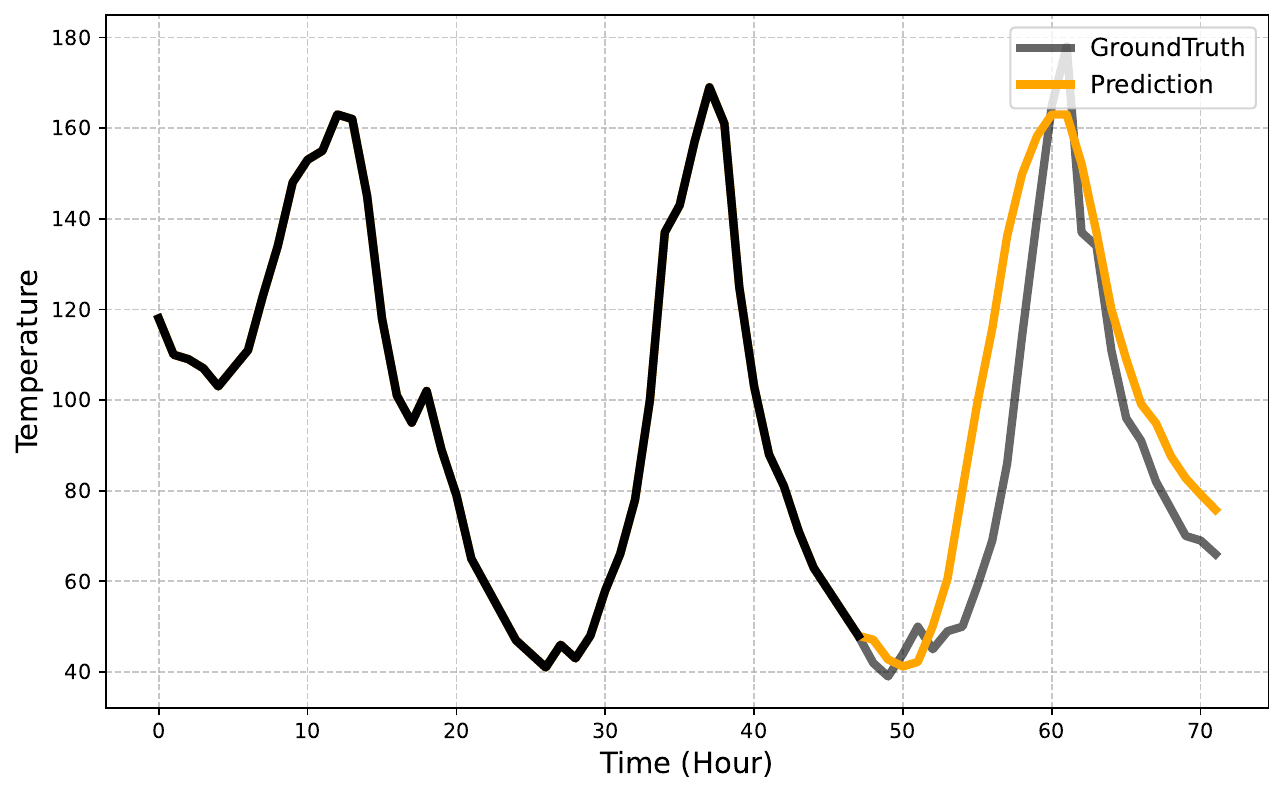}    
}    
\caption{Comparison of single station forecasting results (seasonal sequences). }     
\label{fig:ncei-global-visualization-1}     
\end{figure}

\begin{figure} 
\centering
\subfigure[STID] {     
\includegraphics[width=0.4\columnwidth]{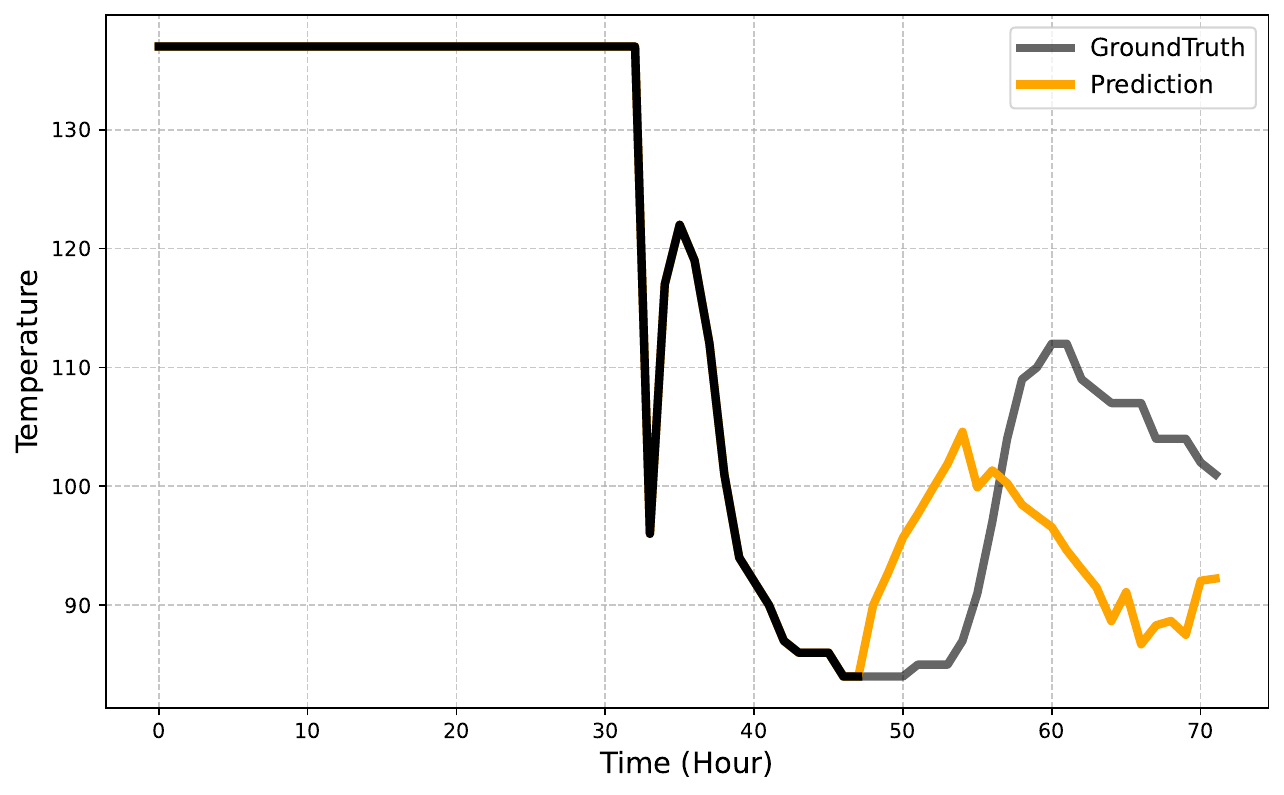}     
}    
\subfigure[RPMixer] {     
\includegraphics[width=0.4\columnwidth]{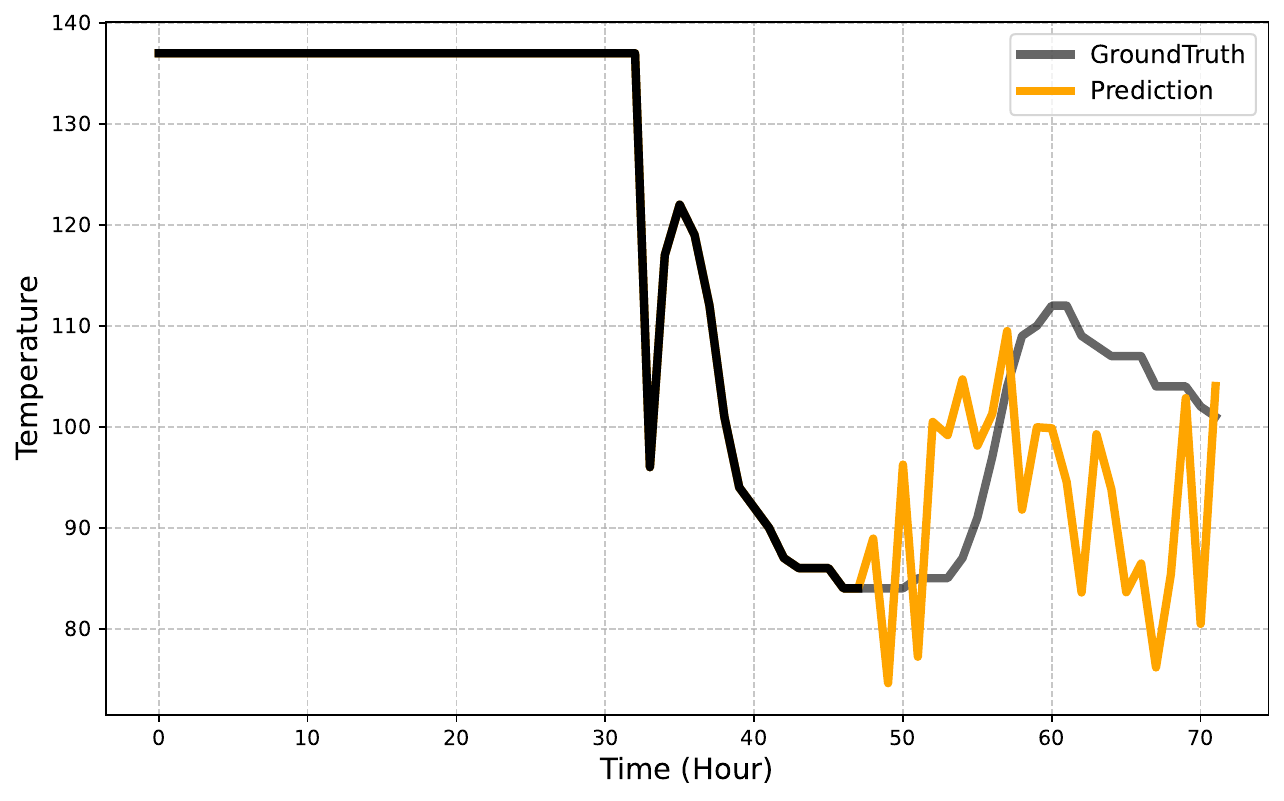}     
}    
\subfigure[PatchSTG] {     
\includegraphics[width=0.4\columnwidth]{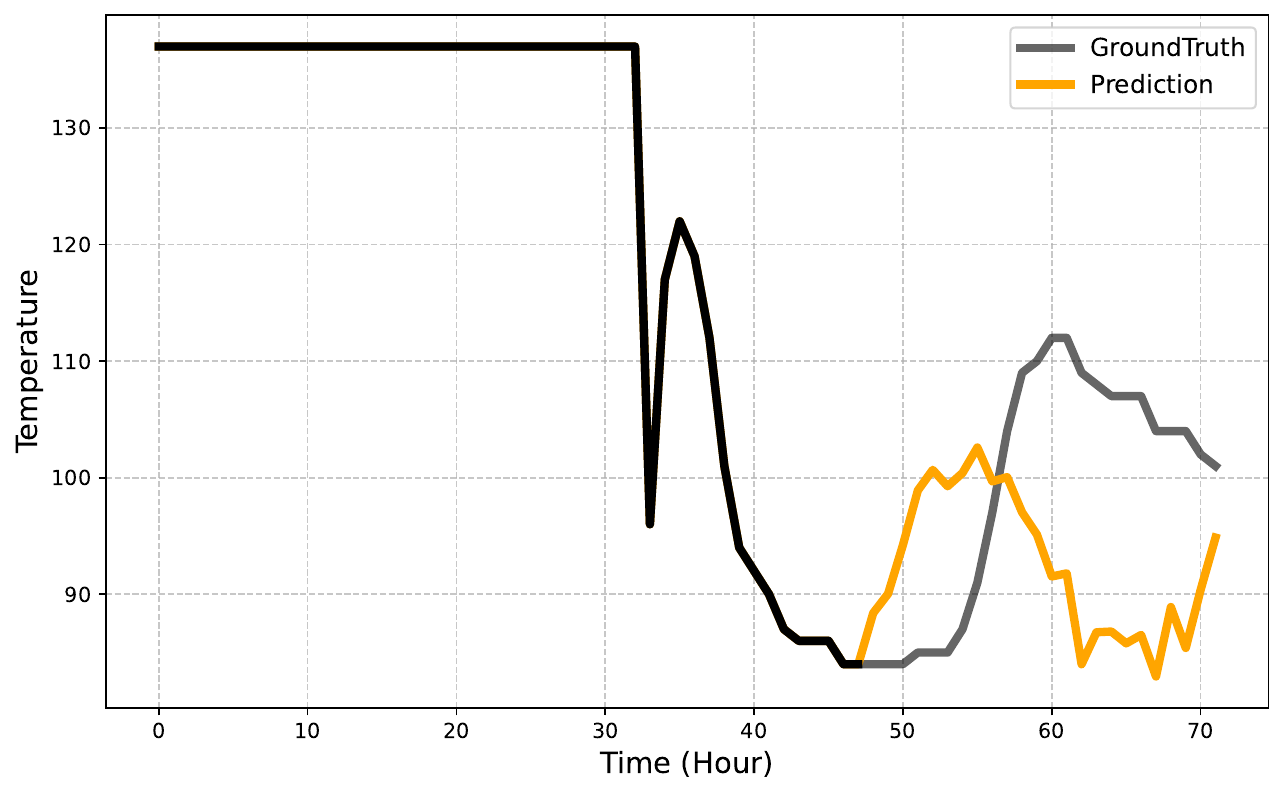}     
}    
\subfigure[S$^2$Transformer] {     
\includegraphics[width=0.4\columnwidth]{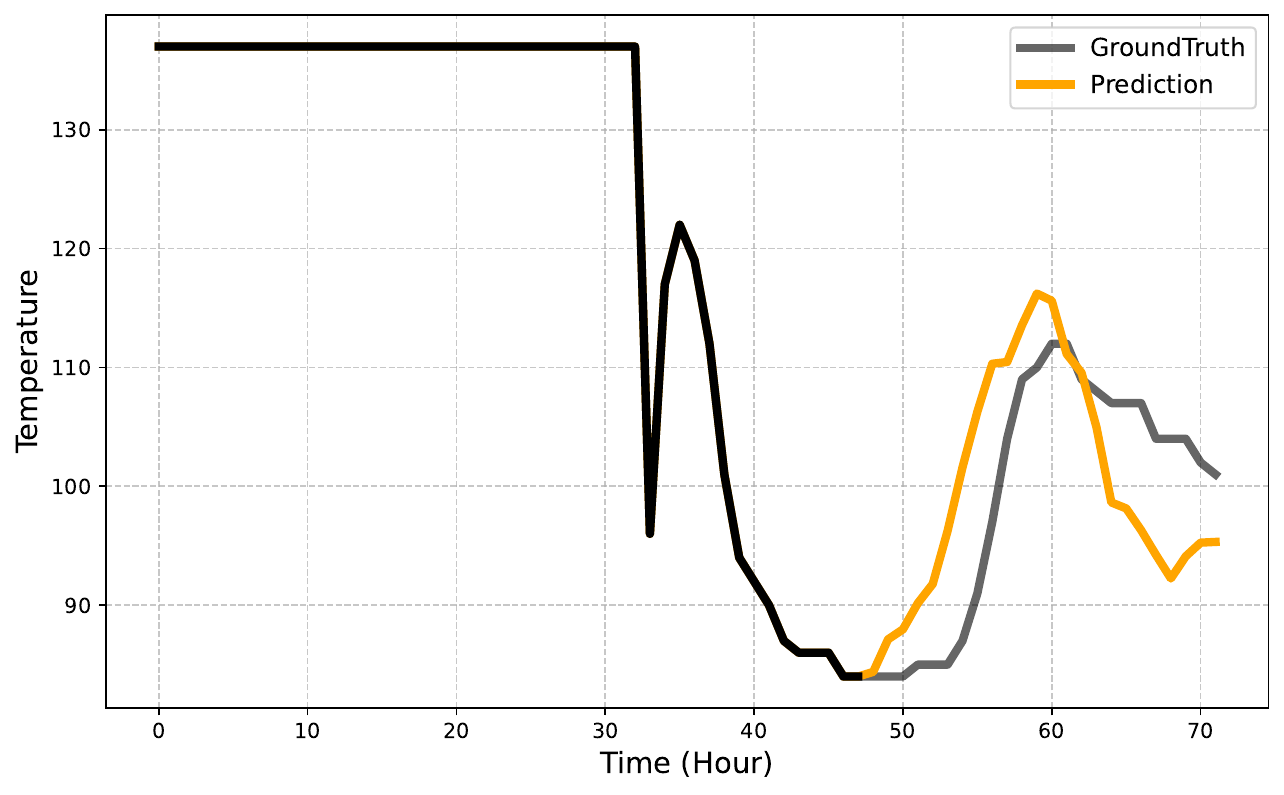}    
}    
\caption{Comparison of single station forecasting results (peak sequences). }     
\label{fig:ncei-global-visualization-2}     
\end{figure}

\textbf{Single Station Forecasting result.} As shown in Figure~\ref{fig:ncei-global-visualization-1} and Figure~\ref{fig:ncei-global-visualization-2}, we plot the predictions of different models for the temperature variable in the NCEI Global Dataset from a single-station perspective. It can be observed that Corrformer outperforms other baseline models in the modeling of seasonal, peak, and stationary sequence values. This is attributed to the model's ability to effectively capture local spatial correlations and leverage changes in neighboring nodes to achieve more accurate forecasting.

\end{document}